\documentclass{article} 

\usepackage[utf8]{inputenc} 
\usepackage[T1]{fontenc}    
\usepackage{hyperref}       
\usepackage{url}            
\usepackage{booktabs}       
\usepackage{amsmath,amsfonts,amsthm}       
\usepackage{amssymb}
\usepackage{mathtools}      
\usepackage{nicefrac}       
\usepackage{microtype}      
\usepackage{subcaption}
\usepackage{mathabx}
\usepackage{bm}
\usepackage{dsfont}
\usepackage{multirow}
\usepackage[inline]{enumitem}
\usepackage{diagbox}
\usepackage{pifont}
\usepackage[capitalise]{cleveref}  
\usepackage{comment}
\usepackage{etoolbox}
\usepackage{xcolor}
\usepackage{graphicx}
\usepackage[ruled,vlined,linesnumbered]{algorithm2e}
\usepackage{algorithmic}
\usepackage[font=small]{caption}
\usepackage{scalerel}
\usepackage{wrapfig}
\usepackage{makecell}
\usepackage{stackengine}

\usepackage[numbers,sort]{natbib}
\newlength{\defbaselineskip}
\newtheorem{definition}{Definition}

\newtheorem{property}{Property}

\allowdisplaybreaks

\title{Multiwavelet-based Operator Learning for Differential Equations}

\usepackage{authblk}
\author{Gaurav Gupta}
\author{Xiongye Xiao}
\author{Paul Bogdan}
\affil{Ming Hsieh Department of Electrical and Computer Engineering,
University of Southern California, Los Angeles, CA 90089}
\affil{{\texttt{\{ggaurav,xiongyex,pbogdan\}@usc.edu}}}
\date{}

\begin{document}

\maketitle

\begin{abstract}
The solution of a partial differential equation can be obtained by computing the inverse operator map between the input and the solution space. Towards this end, we introduce a \textit{multiwavelet-based neural operator learning scheme} that compresses the associated operator's kernel using fine-grained wavelets. By explicitly embedding the inverse multiwavelet filters, we learn the projection of the kernel onto fixed multiwavelet polynomial bases. The projected kernel is trained at multiple scales derived from using repeated computation of multiwavelet transform. This allows learning the complex dependencies at various scales and results in a resolution-independent scheme. Compare to the prior works, we exploit the fundamental properties of the operator's kernel which enable numerically efficient representation. We perform experiments on the Korteweg-de Vries (KdV) equation, Burgers' equation, Darcy Flow, and Navier-Stokes equation. Compared with the existing neural operator approaches, our model shows significantly higher accuracy and achieves state-of-the-art in a range of datasets. For the time-varying equations, the proposed method exhibits a ($2X-10X$) improvement ($0.0018$ ($0.0033$) relative $L2$ error for Burgers' (KdV) equation). By learning the mappings between function spaces, the proposed method has the ability to find the solution of a high-resolution input after learning from lower-resolution data.
\end{abstract}

\allowdisplaybreaks
\section{Introduction}
\label{sec:intro}

Many natural and human-built systems (e.g., aerospace, complex fluids, neuro-glia information processing) exhibit complex dynamics characterized by partial differential equations (PDEs)~\cite{Shlesinger1987,McKeowneaaz2717}. For example, the design of wings and airplanes robust to turbulence, requires to learn complex PDEs. Along the same lines, complex fluids (gels, emulsions) are multiphasic materials characterized by a macroscopic behavior \cite{Ovarlezeaay5589} modeled by non-linear PDEs. Understanding their variations in viscosity as a function of the shear rate is critical for many engineering projects. Moreover, modelling the dynamics of  continuous and discrete cyber and physical processes in complex cyber-physical systems can be achieved through PDEs \cite{YuankunICCPS2016}.

Recent efforts on learning PDEs (i.e., mappings between infinite-dimensional spaces of functions), from trajectories of variables, focused on developing machine learning and in particular deep neural networks (NNs) techniques. Towards this end, a stream of work aims at parameterizing the solution map as deep NNs \cite{Guo2016CNN, Zhu_2018, Bhatnagar_2019, Adler_2017, KHOO_2020}. One issue, however, is that the NNs are tied to a specific resolution during training, and therefore, may not generalize well to other resolutions, thus, requiring retraining (and possible modifications of the model) for every set of discretizations. In parallel, another stream of work focuses on constructing the PDE solution function as a NN architecture \cite{RAISSI2019686, greenfeld2019learning, Kochkove2101784118, wang2021learning}. This approach, however, is designed to work with one instance of a PDE and, therefore, upon changing the coefficients associated with the PDE, the model has to be re-trained. Additionally, the approach is not a complete data-dependent one, and hence, cannot be made oblivious to the knowledge of the underlying PDE structure. Finally, the closest stream of work to the problem we investigate is represented by the ``Neural Operators" \cite{li2020neural, Li2020MGNO, li2020fourier, bhattacharya2020model, PATEL2021113500}. Being a complete data-driven approach, the neural operators method aims at learning the operator map without having knowledge of the underlying PDEs. The neural operators have also demonstrated the capability of discretization-independence. Obtaining the data for learning the operator map could be prohibitively expensive or time consuming (e.g., aircraft performance to different initial conditions). To be able to better solve the problem of learning the PDE operators from scarce and noisy data, we would ideally explore fundamental properties of the operators that have implication
in data-efficient representation.

Our intuition is to transform the problem of learning a PDE to a domain where a compact representation of the operator exists. With a mild assumption regarding the smoothness of the operator's kernel, except finitely many singularities, the multiwavelets \cite{Alpert1993L2Bases}, with their \textit{vanishing moments property}, sparsify the kernel in their projection with respect to (w.r.t.) a measure. Therefore, learning an operator kernel in the multiwavelet domain is feasible and data efficient. The wavelets have a rich history in signal processing \cite{Daubechies1988, Daubechies1992Lectures}, and are popular in audio, image compression~\cite{Balan2009,Silverman1999}. For multiwavelets, the orthogonal polynomial (OP) w.r.t. a measure emerges as a natural basis for the multiwavelet subspace, and an appropriate scale / shift provides a sequence of subspaces which captures the locality at various resolutions. We generalize and exploit the multiwavelets concept to work with arbitrary measures which opens-up new possibilities to design a series of models for the operator learning from complex data streams.

We incorporate the multiwavelet filters derived using a variety of the OP basis into our operator learning model, and show that the proposed architecture outperforms the existing neural operators. Our main contributions are as follows: \textbf{(i)} Based on some fundamental properties of the integral operator's kernel, we develop a multiwavelet-based model which learns the operator map efficiently. \textbf{(ii)} For the 1-D dataset of non-linear Korteweg-de Vries and Burgers equations, we observe an order of magnitude improvement in the relative $L2$ error (Section\,\ref{ssec:kdv}, \ref{ssec:burgers}). \textbf{(iii)} We demonstrate that the proposed model is in validation with the theoretical properties of the pseudo-differential operator (Section\,\ref{ssec:eulerBer}). \textbf{(iv)} We show how the proposed multiwavelet-based model is robust towards the fluctuation strength of the input signal (Section\,\ref{ssec:kdv}). \textbf{(v)} Next, we demonstrate the applicability on higher dimensions of 2-D Darcy flow equation (Section\,\ref{ssec:darcy}), and finally show that the proposed approach can learn at lower resolutions and generalize to higher resolutions. The code for reproducing the experiments is available at: \url{https://github.com/gaurav71531/mwt-operator}.

\section{Operator Learning using Multiwavelet Transform}
\label{sec:opLearning}
We start by defining the problem of operator learning in Section\,\ref{ssec:probSet}.
Section\,\ref{ssec:nsMWT} defines the multiwavelet transform for the proposed operator learning problem and derives the necessary transformation operations across different scales. Section\,\ref{ssec:mwtModel} outlines the proposed operator learning model. Finally, Section\,\ref{ssec:OpProp} lists some of the useful properties of the operators which leads to an efficient implementation of multiwavelet-based models.
\subsection{Problem Setup}
\label{ssec:probSet}
Given two functions $a(x)$ and $u(x)$ with $x\in D$, the operator is a map $T$ such that $Ta = u$. Formally, let $\mathcal{A}$ and $\mathcal{U}$ be two Sobolev spaces $\mathcal{H}^{s,p}$ $(s>0, p\geq 1)$, then the operator $T$ is such that $T:\mathcal{A}\rightarrow\mathcal{U}$. The Sobolev spaces are particularly useful in the analysis of partial differential equations (PDEs), and we restrict our attention to $s>0$ and $p=2$. Note that, for $s=0$, the $\mathcal{H}^{0,p}$ coincides with $L^{p}$, and, $f\in \mathcal{H}^{0,p}$ does not necessarily have derivatives in $L^{p}$. We choose $p=2$ in order to be able to define projections with respect to (w.r.t.) measures $\mu$ in a Hilbert space structure.

We take the operator $T$ as an integral operator with the kernel $K:D\times D\rightarrow L^{2}$ such that
\begin{equation}
    Ta(x) = \int\nolimits_{D} K(x, y)a(y)dy.
    \label{eq:kernelInt}
\end{equation}
For the case of inhomogeneous linear PDEs, $\mathcal{L}u = f$, with $f$ being the forcing function, $\mathcal{L}$ is the differential operator, and the associated kernel is commonly termed as Green function. In our case, we do not put the restriction of linearity on the operator. From eq. \eqref{eq:kernelInt}, it is apparent that learning the complete kernel $K(.,.)$ would essentially solve the operator map problem, but it is not necessarily a numerically feasible solution. Indeed, a better approach would be to exploit possible useful properties (see Section\,\ref{ssec:OpProp}) such that a compact representation of the kernel can be made. For an efficient representation of the operator kernel, we need an appropriate subspace (or sequence of subspaces), and projection tools to map to such spaces.

\textbf{Norm with respect to measures:} Projecting a given function onto a fixed basis would require a measure dependent distance. For two functions $f$ and $g$, we take the inner product w.r.t measure $\mu$ as $\langle f, g\rangle_{\mu} = \int f(x)g(x)d\mu(x)$, and the associated norm as $\vert\vert f\vert\vert_{\mu} = \langle f, f\rangle_{\mu}^{1/2}$. We now discuss the next ingredient, which refers to the subspaces required to project the kernel.

\subsection{Multiwavelet Transform}
\label{ssec:nsMWT}
In this section, we briefly overview the concept of multiwavelets \cite{ALPERT2002MWT} and extend it to work with non-uniform measures at each scale. The multiwavelet transform synergizes the advantages of \textit{orthogonal polynomials} (OPs) as well as the \textit{wavelets} concepts, both of which have a rich history in the signal processing. The properties of wavelet bases like (\textit{i}) vanishing moments, and (\textit{ii}) orthogonality can effectively be used to create a system of coordinates in which a wide class of operators (see Section\,\ref{ssec:OpProp}) have \textit{nice} representation. Multiwavelets go few steps further, and provide a fine-grained representation using OPs, but also act as basis on a finite interval. For the rest of this section, we restrict our attention to the interval $[0, 1]$; however, the transformation to any finite interval $[a,b]$ could be straightforwardly obtained by an appropriate shift and scale.

\textbf{Multi Resolution Analysis:} We begin by defining the space of piecewise polynomial functions, for $k\in\mathbb{N}$ and $n\in\mathbb{Z}^{+}\cup\{0\}$ as, ${\bf{V}}_{n}^{k} = \bigcup\nolimits_{l=0}^{2^n-1} \{f \vert \text{deg}(f)<k\,\, \text{for}\,\, x\in (2^{-n}l, 2^{-n}(l+1))\wedge 0,\,\, \text{elsewhere} \}$. Clearly, $\text{dim}({\bf V}_{n}^{k}) = 2^{n}k$, and for subsequent $n$, each subspace is contained in another as shown by the following relation:
\begin{equation}
    {\bf V}_{0}^{k} \subset {\bf V}_{1}^{k} \hdots \subset {\bf V}_{n-1}^{k} \subset {\bf V}_{n}^{k} \subset \hdots.
    \label{eqn:vspace}
\end{equation}
Similarly, we define the sequence of measures $\mu_{0}, \mu_{1}, \hdots$ such that $f\in{\bf V}_{n}^{k}$ is measurable w.r.t. $\mu_{n}$ and the norm of $f$ is taken as $\vert\vert f\vert\vert = \langle f, f\rangle_{\mu_{n}}^{1/2}$. Next, since ${\bf V}_{n-1}^{k} \subset {\bf V}_{n}^{k}$, we define the multiwavelet subspace as ${\bf W}_{n}^{k}$ for $n\in\mathbb{Z}^{+}\cup\{0\}$, such that 
\begin{equation}
    {\bf V}_{n+1}^{k} = {\bf V}_{n}^{k}\bigoplus{\bf W}_{n}^{k}, \quad {\bf V}_{n}^{k} \perp {\bf W}_{n}^{k}.
    \label{eqn:waveletSpace}
\end{equation}
For a given OP basis for ${\bf V}_{0}^{k}$ as $\phi_{0}, \phi_{1}, \hdots, \phi_{k-1}$ w.r.t. measure $\mu_{0}$, a basis of the subsequent spaces ${\bf V}_{n}^{k}, n>1$ can be obtained by shift and scale (hence the name, multi-scale) operations of the original basis as follows:
\begin{equation}
    \phi_{jl}^{n}(x) = 2^{n/2}\phi_{j}(2^{n}x-l),\quad j = 0, 1, \hdots, k-1, \quad l = 0, 1, \hdots, 2^n-1, \text{w.r.t.}\quad \mu_{n},
    \label{eqn:scaleHigh}
\end{equation}
\noindent where, $\mu_{n}$ is obtained as the collections of shift and scale of $\mu_{0}$, accordingly.

\textbf{Multiwavelets:} For the multiwavelet subspace ${\bf W}_{0}^{k}$, the orthonormal basis (of piecewise polynomials) are taken as $\psi_{0}, \psi_{1}, \hdots, \psi_{k-1}$ such that $\langle \psi_{i}, \psi_{j}\rangle_{\mu_{0}} = 0$ for $i\neq j$ and $1$, otherwise. From eq. \eqref{eqn:waveletSpace}, ${\bf V}_{n}^{k} \perp {\bf W}_{n}^{k}$, and since ${\bf V}_{n}^{k}$ spans the polynomials of degree at most $k$, therefore, we conclude that
\begin{equation}
    \int\limits_{0}^{1}x^{i}\psi_{j}(x)d\mu_{0}(x) = 0,\quad \forall\, 0\leq j,i <k. \qquad \text{(vanishing moments)}
    \label{eqn:vanishingMom}
\end{equation}
Similarly to eq. \eqref{eqn:scaleHigh}, a basis for multiwavelet subspace ${\bf W}_{n}^{k}$ are obtained by shift and scale of $\psi_{i}$ as $\psi_{jl}^{n}(x) = 2^{n/2}\psi_{j}(2^nx-l)$ and $\psi_{jl}^{n}$ are orthonormal w.r.t. measure $\mu_{n}$, i.e. $\langle \psi_{jl}^{n}, \psi_{j^{\prime}l^{\prime}}^{n}\rangle_{\mu_{n}} = 1$ if $j=j^\prime, l=l^\prime$, and $0$ otherwise. Therefore, for a given OP basis for ${\bf V}_{0}^{k}$ (for example, Legendre, Chebyshev polynomials), we only require to compute $\psi_{i}$, and a complete basis set at all the scales can be obtained using scale/shift of $\phi_{i}, \psi_{i}$.

\textbf{Note:} Since ${\bf V}_{1}^{k} = {\bf V}_{0}^{k}\bigoplus{\bf W}_{0}^{k}$ from eq. \eqref{eqn:waveletSpace}, therefore, for a given basis $\phi_{i}$ of ${\bf V}_{0}^{k}$ w.r.t. measure $\mu_{0}$ and $\phi_{jl}^{n}$ as a basis for ${\bf V}_{1}^{k}$ w.r.t. $\mu_{1}$, a set of basis $\psi_{i}$ can be obtained by applying Gram-Schmidt Orthogonalization using appropriate measures. We refer the reader to supplementary materials for the detailed procedure.

\textbf{Note:} Since ${\bf V}_{0}^{k}$ and ${\bf W}_{0}^{k}$ lives in ${\bf V}_{1}^{k}$, therefore, $\phi_{i}, \psi_{i}$ can be written as a linear combination of the basis of $V_{1}^{k}$. We term these linear coefficients as multiwavelet decomposition filters $(H^{(0)}, H^{(1)}, G^{(0)}, G^{(1)})$, since they are transforming a fine $n=1$ to coarse scale $n=0$. A uniform measure ($\mu_{0}$) version is discussed in \cite{ALPERT2002MWT}, and we extend it to any arbitrary measure by including the correction terms $\Sigma^{(0)}$ and $\Sigma^{(1)}$. We refer to supplementary materials for the complete details. The capability of using the non-uniform measures enables us to apply the same approach to any OP basis with finite domain, for example, Chebyshev, Gegenbauer, etc.

For a given $f(x)$, the multiscale, multiwavelet coefficients at the scale $n$ are defined as ${\bf s}_{l}^{n} = [\langle f, \phi_{il}^{n}\rangle_{\mu_{n}}]_{i=0}^{k-1}$, ${\bf d}_{l}^{n} = [\langle f, \psi_{il}^{n}\rangle_{\mu_{n}}]_{i=0}^{k-1}$, respectively, w.r.t. measure $\mu_{n}$ with ${\bf s}_{l}^{n}, {\bf d}_{l}^{n}\in\mathbb{R}^{k\times 2^{n}}$. The decomposition / reconstruction across scales is written as

\begin{minipage}{0.49\linewidth}
\begin{align}
    {\bf s}_{l}^{n} = H^{(0)}{\bf s}_{2l}^{n+1} + H^{(1)}{\bf 
    s}_{2l+1}^{n+1},\label{eqn:scaleDecFilter}\\
    {\bf d}_{l}^{n} = G^{(0)}{\bf s}_{2l}^{n+1} + H^{(1)}{\bf s}_{2l+1}^{n+1}.\label{eqn:wavelDecFilter}
\end{align}
\end{minipage}
\hfill
\begin{minipage}{0.49\linewidth}
\begin{align}
    {\bf s}_{2l}^{n+1} = \Sigma^{(0)}(H^{(0)\,T}{\bf s}_{l}^{n} + G^{(0)\,T}{\bf d}_{l}^{n}),\label{eqn:evenScaleRec}\\
    {\bf s}_{2l+1}^{n+1} = \Sigma^{(1)}(H^{(1)\,T}{\bf s}_{l}^{n} + G^{(1)\,T}{\bf d}_{l}^{n}).\label{eqn:OddScaleRec}
\end{align}
\end{minipage}

The wavelet (and also multiwavelet) transformation can be straightforwardly extended to multiple dimensions using tensor product of the bases. For our purpose, a function $f\in\mathbb{R}^{d}$ has multiscale, multiwavelet coefficients ${\bf s}_{l}^{n}, {\bf d}_{l}^{n} \in \mathbb{R}^{k\times\hdots\times k\times2^{n}}$ which are also recursively obtained by replacing the filters in eq. \eqref{eqn:scaleDecFilter}-\eqref{eqn:wavelDecFilter} with their Kronecker product, specifically, $H^{(0)}$ with $H^{(0)}\otimes H^{(0)}\otimes\hdots H^{(0)}$, where $\otimes$ is the Kronecker product repeated $d$ times. For eq. \eqref{eqn:evenScaleRec}-\eqref{eqn:OddScaleRec} $H^{(0)}\Sigma^{(0)}$ (and similarly others) are replaced with their $d$-times Kronecker product.

\begin{figure*}
	\centering
	\includegraphics[width=\linewidth]{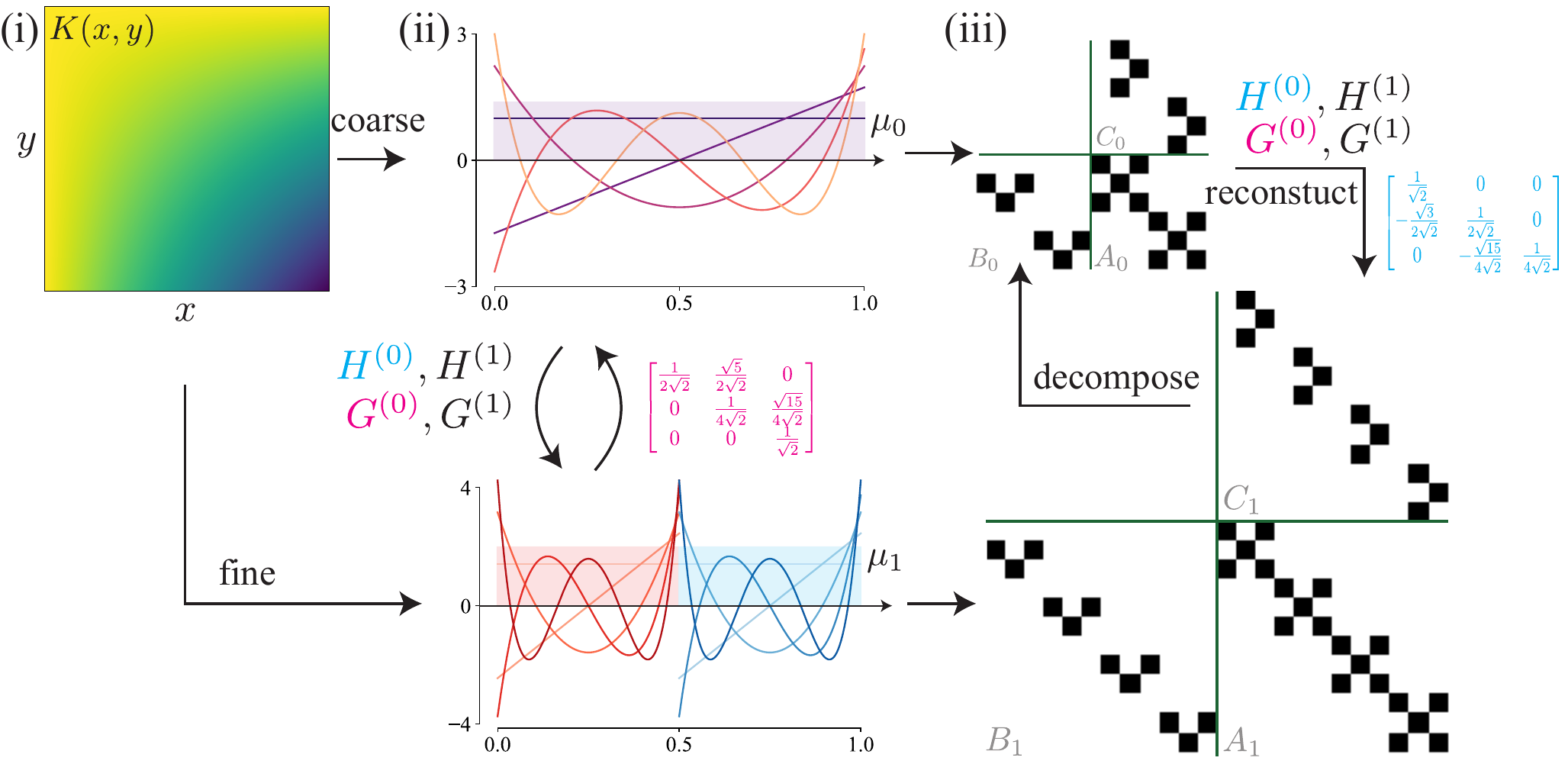}
	\caption{\textbf{Multiwavelet representation of the Kernel}. (i) Given kernel $K(x,y)$ of an integral operator $T$, (ii) the bases with different measures ($\mu_{0}, \mu_{1}$) at two different scales (coarse=$0$, fine=$1$) projects the kernel into 3 components $A_{i}, B_{i}, C_{i}$. (iii) The decomposition yields a sparse structure, and the entries with absolute magnitude values exceeding $1e^{-8}$ are shown in black. Given projections at any scale, the finer / coarser scale projections can be obtained by reconstruction / decomposition using a fixed multiwavelet filters $H^{(i)}$ and $G^{(i)}, i=0,1$.}
	\label{fig:mwtKernel}
\end{figure*}
\textbf{Non-Standard Form:} The multiwavelet representation of the operator kernel $K(x,y)$ can be obtained by an appropriate tensor product of the multiscale and multiwavelet basis. One issue, however, in this approach, is that the basis at various scales are \textit{coupled} because of the tensor product. To untangle the basis at various scales, we use a trick as proposed in \cite{Belykin1991FastWavelet} called the non-standard wavelet representation. The extra mathematical price paid for the non-standard representation, actually serves as a ground for reducing the proposed model complexity (see Section\,\ref{ssec:mwtModel}), thus, providing data efficiency. For the operator under consideration $T$ with integral kernel $K(x, y)$, let us denote $T_{n}$ as the projection of $T$ on $V_{n}^{k}$, which essentially is obtained by projecting the kernel $K$ onto basis $\phi_{jl}^{n}$ w.r.t. measure $\mu_{n}$. If $P_{n}$ is the projection operator such that $P_{n}f = \sum\nolimits_{j,l}\langle f, \phi_{jl}^{n}\rangle_{\mu_{n}}\phi_{jl}^{n}$, then $T_{n} = P_{n}TP_{n}$. Using telescopic sum, $T_{n}$ is expanded as 
\begin{equation}
    T_{n} = \sum\nolimits_{i=L+1}^{n}(Q_{i}TQ_{i} + Q_{i}TP_{i-1} + P_{i-1}TQ_{i}) + P_{L}TP_{L},
\end{equation}
where, $Q_{i} = P_{i}-P_{i-1}$ and $L$ is the coarsest scale under consideration $(L\geq 0)$. From eq. \eqref{eqn:waveletSpace}, it is apparent that $Q_{i}$ is the multiwavelet operator. Next, we denote $A_{i} = Q_{i}TQ_{i}, B_{i} = Q_{i}TP_{i-1}, C_{i} = P_{i-1}TQ_{i}$, and $\bar{T} = P_{L}TP_{L}$. In Figure\,\ref{fig:mwtKernel}, we show the non-standard multiwavelet transform for a given kernel $K(x, y)$. The transformation has a sparse banded structure due to smoothness property of the kernel (see Section\,\ref{ssec:OpProp}). For the operator $T$ such that $Ta = u$, the map under multiwavelet domain is written as
\begin{align}
\begin{aligned}
U_{d\,l}^{n} &=  A_{n}d_{l}^{n} + B_{n}s_{l}^{n}, \qquad U_{\hat{s}\,l}^{n} = C_{n}d_{l}^{n}, \qquad U_{s\,l}^{L} = \bar{T}s_{l}^{L},
\label{eqn:kernelNSMWT}
\end{aligned}
\end{align}
where, $(U_{s\,l}^{n}, U_{d\,l}^{n})/(s_{l}^{n}, d_{l}^{n})$ are the multiscale, multiwavelet coefficients of $u / a$, respectively, and $L$ is the coarsest scale under consideration. With these mathematical concepts, we now proceed to define our multiwavelet-based operator learning model in the Section\,\ref{ssec:mwtModel}.

\begin{figure*}
	\centering
	\includegraphics[width=0.9\linewidth]{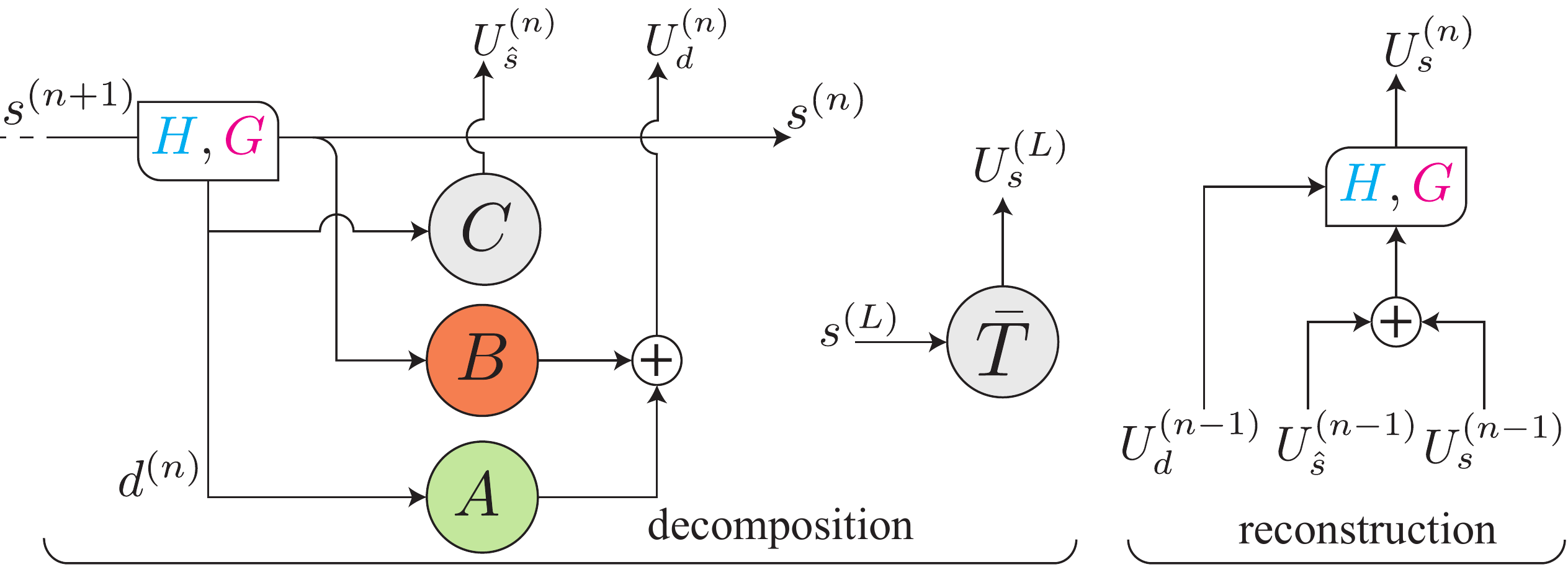}
	\caption{\textbf{MWT model architecture}. (\textbf{Left)} Decomposition cell using 4 neural networks (NNs) $A, B$ and $C$, and $T$ (for the coarsest scale $L$) performs multiwavelet decomposition from scale $n+1$ to $n$. \textbf{(Right)} Reconstruction module using pre-defined filters $H^{(i)}, G^{(i)}$ performs inverse multiwavelet transform from scale $n-1$ to $n$.}
	\label{fig:mwtModel}
	\vspace{-10pt}
\end{figure*}
\subsection{Multiwavelet-based Model}
\label{ssec:mwtModel}
Based on the discussion in Section\,\ref{ssec:nsMWT}, we propose a multiwavelet-based model (MWT) as shown in Figure\,\ref{fig:mwtModel}. For a given input/output as $a/u$, the goal of the MWT model is to map the multiwavelet-transform of the input $(s_{l}^{N})$ to output $(U_{s\,l}^{N})$ at the finest scale $N$. The model consists of two parts: (\textit{i}) Decomposition (\textit{dec}), and (\textit{ii}) Reconstruction (\textit{rec}). The \textit{dec} acts as a recurrent network, and at each iteration the input is $s^{n+1}$. Using \eqref{eqn:scaleDecFilter}-\eqref{eqn:wavelDecFilter}, the input is used to obtain multiscale and multiwavelet coefficients at a coarser level $s^{n}$ and $d^{n}$, respectively. Next, to compute the multiscale/multiwavelet coefficients of the output $u$, we approximate the non-standard kernel decomposition from \eqref{eqn:kernelNSMWT} using four neural networks (NNs) $A, B, C$ and $\bar{T}$ such that $U_{d\,l}^{n}\approx A_{\theta_{A}}(d_{l}^{n})+ B_{\theta_{B}}(s_{l}^{n}), U_{\hat{s}\,l}^{n}\approx C_{\theta_{C}}(d_{l}^{n}), \forall\, 0\leq n<L$, and $U_{s\,l}^{L}\approx \bar{T}_{\theta_{\bar{T}}}(s_{l}^{L})$. This is a ladder-down approach, and the \textit{dec} part performs the decimation of signal (factor $1/2$), running for a maximum of $L$ cycles, $L<\log_{2}(M)$ for a given input sequence of size $M$. Finally, the \textit{rec} module collects the constituent terms $U_{s\,l}^{n}, U_{\hat{s}\,l}^{n}, U_{d\,l}^{n}$ (obtained using the \textit{dec} module) and performs a ladder-up operation to compute the multiscale coefficients of the output at a finer scale $n+1$ using \eqref{eqn:evenScaleRec}-\eqref{eqn:OddScaleRec}. The iterations continue until the finest scale $N$ is obtained for the output.

At each iteration, the filters in \textit{dec} module downsamples the input, but compared to popular techniques (e.g., maxpool), the input is only transformed to a coarser multiscale/multiwavelet space. By virtue of its design, since the non-standard wavelet representation does not have inter-scale interactions, it basically allows us to reuse the same kernel NNs $A, B, C$ at different scales. A follow-up advantage of this approach is that the model is resolution independent, since the recurrent structure of \textit{dec} is input invariant, and for a different input size $M$, only the number of iterations would possibly change for a maximum of $\log_{2}M$. The reuse of $A, B, C$ by re-training at various scales also enable us to learn an expressive model with fewer parameters $(\theta_{A}, \theta_{B}, \theta_{C}, \theta_{\bar{T}})$. We see in Section\,\ref{sec:empEval}, that even a single-layered CNN for $A, B, C$ is sufficient for learning the operator.

The \textit{dec} / \textit{rec} module uses the filter matrices which are fixed beforehand, therefore, this part does not require any training. The model does not work for any arbitrary choice of fixed matrices $H, G$. We show in Section\,\ref{ssec:darcy} that for randomly selected matrices, the model does not learn, which validates that careful construction of filter matrices is necessary.

\subsection{Operators Properties}
\label{ssec:OpProp}
This section outlines definition of the integral kernels that are typically useful in an efficient compression of the operators through multiwavelets. We then discuss a fundamental property of the pseudo-differential operator.
\begin{definition}[\cite{meyer1997wavelets}]
\textbf{Calder\'on-Zygmund Operator}. The integral operators that have kernel $K(x, y)$ which is smooth away from the diagonal, and satisfy the following.
\begin{align}
\begin{aligned}
\vert K(x, y)\vert &\leq \frac{1}{\vert x-y\vert},\\
\vert \partial_{x}^{M} K(x, y)\vert + \vert \partial_{y}^{M} K(x, y)\vert &\leq \frac{C_{0}}{\vert x - y\vert^{M+1}}.
\end{aligned}
\end{align}
\label{def:CZ}
\end{definition}
The smooth functions with decaying derivatives are \textit{gold} to the multiwavelet transform. Note that, smoothness implies Taylor series expansion, and the multiwavelet transform with sufficiently large $k$ zeroes out the initial $k$ terms of the expansion due to vanishing moments property \eqref{eqn:vanishingMom}. This is how multiwavelet sparsifies the kernel (see Figure\,\ref{fig:mwtKernel} where $K(x,y)$ is smooth). Although, the definition of Calder\'on-Zygmund is simple (singularities only at the diagonal), but the multiwavelets are capable to compresses the kernel as long as the \textit{number of singularities are finite}.

The next property, from \cite{Chou2000Green}, points out that with input/output being single-dimensional functions, for any pseudo-differential operator (with smooth coefficients), the singularity at the diagonal is also well-characterized.
\begin{property}
\textbf{Smoothness of Pseudo-Differential Operator.} For the integral kernel K(x,y) of a pseudo-differential operator, $K(x,y)\in C^{\infty}$  $\forall x\neq y$, and for $x=y$, $K(x,y) \in C^{T-1}$, where $T+1$ is the highest derivative order in the given pseudo-differential equation.
\label{propr:smoothness}
\end{property}
The property\,\ref{propr:smoothness} implies that, for the class of pseudo-differential operator, and any set of basis with the initial $J$ vanishing moments, the projection of kernel onto such bases will have the diagonal dominating the non-diagonal entries, exponentially, if $J>T-1$ \cite{Chou2000Green}. For the case of multiwavelet basis with $k$ OPs, $J=k$ (from eq. \eqref{eqn:vanishingMom}). Therefore, $k>T-1$ sparsifies the kernel projection onto multiwavelets, for a fixed number of bits precision $\epsilon$. We see the implication of the Property\,\ref{propr:smoothness} on our proposed model in the Section\,\ref{ssec:eulerBer}.

\section{Empirical Evaluation}
\label{sec:empEval}
In this section, we evaluate the multiwavelet-based model (MWT) on several PDE datasets. We show that the proposed MWT model not only exhibits orders of magnitude higher accuracy when compared against the state-of-the-art (Sota) approaches but also works consistently well under different input conditions without parameter tuning. From a numerical perspective, we take the data as point-wise evaluations of the input and output functions. Specifically, we have the dataset $(a_{i}, u_{i})$ with $a_{i} = a(x_{i}), u_{i} = u(x_{i})$ for  $x_{1}, x_{2},\hdots, x_{N}\in D$, where $x_{i}$ are $M$-point discretization of the domain $D$. Unless stated otherwise, the training set is of size $1000$ while test is of size $200$.

\textbf{Model architectures:} Unless otherwise stated, the NNs $A, B$ and $C$ in the proposed model (Figure\,\ref{fig:mwtModel}) are chosen as a single-layered CNNs following a linear layer, while $\bar{T}$ is taken as single $k\times k$ linear layer. We choose $k=4$ in all our experiments, and the OP basis as Legendre (Leg), Chebyshev (Chb) with uniform, non-uniform measure $\mu_{0}$, respectively. The model in Figure\,\ref{fig:mwtModel} is treated as single layer, and for $1$D equations, we cascade $2$ multiwavelet layers, while for $2$D dataset, we use a total $4$ layers with $ReLU$ non-linearity.

From a mathematical viewpoint, the \textit{dec} and \textit{rec} modules in Figure\,\ref{fig:mwtModel} transform only the multiscale and multiwavelet coefficients. However, the input and output to the model are point-wise function samples, i.e., $(a_i, u_i)$. A remedy around this is to take the data sequence, and construct hypothetical functions $f_{a} = \sum\nolimits_{i=1}^{N}a_{i}\phi_{ji}^{n}$ and $f_{u} = \sum\nolimits_{i=1}^{N}u_{i}\phi_{ji}^{n}$. Clearly, $f_{a}, f_{u}$ lives in $V_{n}^{k}$ with $n = \log_{2}N$. Now the model can be used with $s^{(n)} = a_{i}$ and $U_{s}^{(n)} = u_i$. Note that $f_a, f_u$ are not explicitly used, but only a matter of convention.

\textbf{Benchmark models:} We compare our \textbf{MWT} model using two different OP basis (Leg, Chb) with the most recent successful neural operators. Specifically, we consider the graph neural operator (\textbf{GNO}) \cite{li2020neural}, the multipole graph neural operator (\textbf{MGNO}) \cite{Li2020MGNO}, the \textbf{LNO} which makes a low-rank ($r$) representation of the operator kernel $K(x,y)$ (also similar to unstacked DeepONet \cite{lu2020deeponet}),  and the Fourier neural operator (\textbf{FNO} ) \cite{li2020fourier}. We experiment on three competent datasets setup by the work of FNO (Burgers' equation (1-D), Darcy Flow (2-D), and Navier-Stokes equation (time-varying 2-D)). In addition, we also experiment with Korteweg-de Vries equation (1-D). For the 1-D cases, a modified FNO with careful parameter selection and removal of Batch-normalization layers results in a better performance compared with the original FNO, and we use it in our experiments. The MWT model demonstrates the highest accuracy in all the experiments. The MWT model also shows the ability to learn the function mapping through lower-resolution data, and able to generalize to higher resolutions.

All the models (including ours) are trained for a total of $500$ epochs using Adam optimizer with an initial learning rate (LR) of $0.001$. The LR decays after every $100$ epochs with a factor of $\gamma=0.5$. The loss function is taken as relative $L2$ error \cite{li2020fourier}. All of the experiments are performed on a single Nvidia V100 32 GB GPU, and the results are averaged over a total of $3$ seeds.

\subsection{Korteweg-de Vries (KdV) Equation}
\label{ssec:kdv}
\begin{table}
\centering
\begin{tabular}{p{2cm} p{1.8cm} p{1.8cm} p{1.8cm} p{1.8cm} p{1.8cm}}
 \hline
 Networks& s = 64&s = 128&s = 256&s = 512 &s = 1024\\
 \hline
 MWT Leg    &\textbf{0.00338}&\textbf{0.00375}&\textbf{0.00418}&\textbf{0.00393}&\textbf{0.00389}\\
 MWT Cheb &0.00715&0.00712&0.00604&0.00769&0.00675 \\
 \hline
FNO & 0.0125 &  0.0124 &  0.0125 &  0.0122 & 0.0126\\
 MGNO& 0.1296    &0.1515&0.1355&0.1345&0.1363\\
LNO&   0.0429&0.0557&0.0414&0.0425 & 0.0447\\
GNO & 0.0789 & 0.0760 & 0.0695 & 0.0699 & 0.0721\\
 \hline 
\end{tabular}
\caption{\label{tab:kdv} Korteweg-de Vries (KdV) equation benchmarks for different input resolution $s$. Top: Our methods. Bottom: previous works of Neural operator.}
\end{table}
The Korteweg-de Vries (KdV) equation was first proposed by Boussinesq~\cite{boussinesq1877essai} and rediscovered by Korteweg and de Vries~\cite{darrigol2005worlds}. KdV is a 1-D non-linear PDE commonly used to describe the non-linear shallow water waves. For a given field $u(x,t)$, the dynamics takes the following form:
\begin{align}
    \begin{aligned}
    \frac{\partial u}{\partial t}&=-0.5u\frac{\partial u}{\partial x}-\frac{\partial^3 u}{\partial x^3}, x\in(0,1), t\in(0,1]\\
    u_{0}(x)&=u(x,t=0)    
    \end{aligned}
    \label{eqn:kdv}
\end{align}
The task for the neural operator is to learn the mapping of the initial condition $u_{0}(x)$ to the solutions $u(x,t=1)$. We generate the initial condition in Gaussian random fields according to $u_{0}\sim \mathcal N(0,7^4(-\Delta+7^2I)^{-2.5})$ with periodic boundary conditions. The equation is numerically solved using \textit{chebfun} package \cite{Driscoll2014} with a resolution $2^{10}$, and datasets with lower resolutions are obtained by sub-sampling the highest resolution data set.

\begin{figure}
    \centering
    \includegraphics[width = \linewidth]{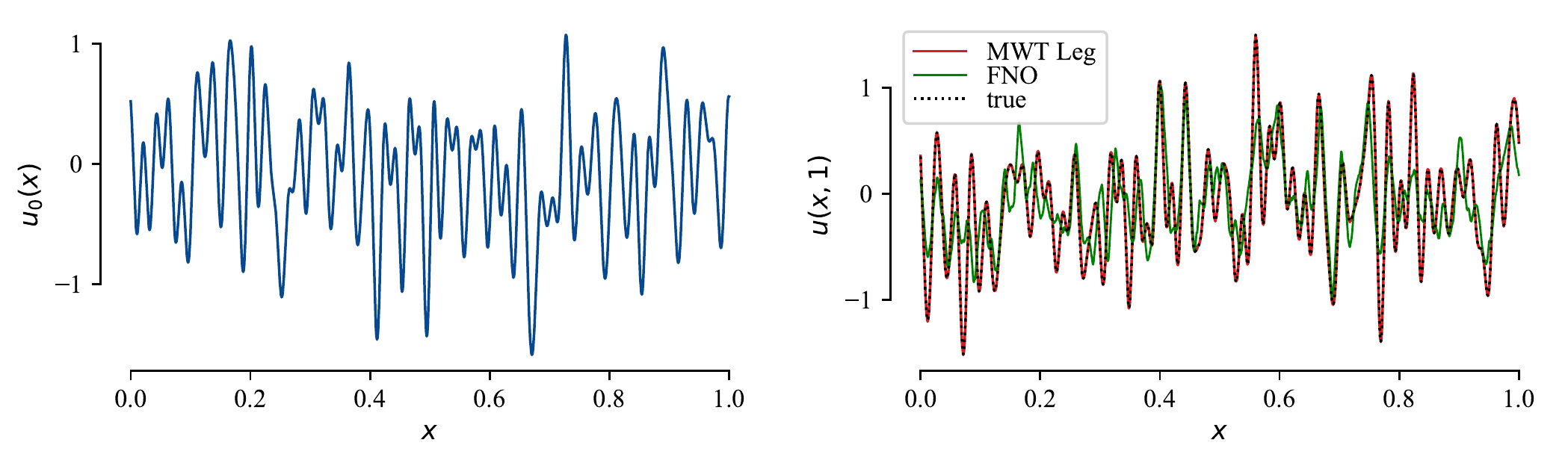}
    \caption{\textbf{The output of the KdV equation}. (Left) An input $u_{0}(x)$ with $\lambda = 0.02$. (Right) The predicted output of the MWT Leg model learning the high fluctuations.}
    \label{fig:KdV_fluctuations}
\end{figure}
\textbf{Varying resolution:} The experimental results of the KdV equation for different input resolutions $s$ are shown in Table\ref{tab:kdv}. We see that, compared to any of the benchmarks, our proposed MWT Leg exhibits the lowest relative error and is lowest nearly by an order of magnitude. Even in the case of the resolution of $64$, 
\begin{wrapfigure}{r}{0.45\linewidth}
\includegraphics[width=\linewidth]{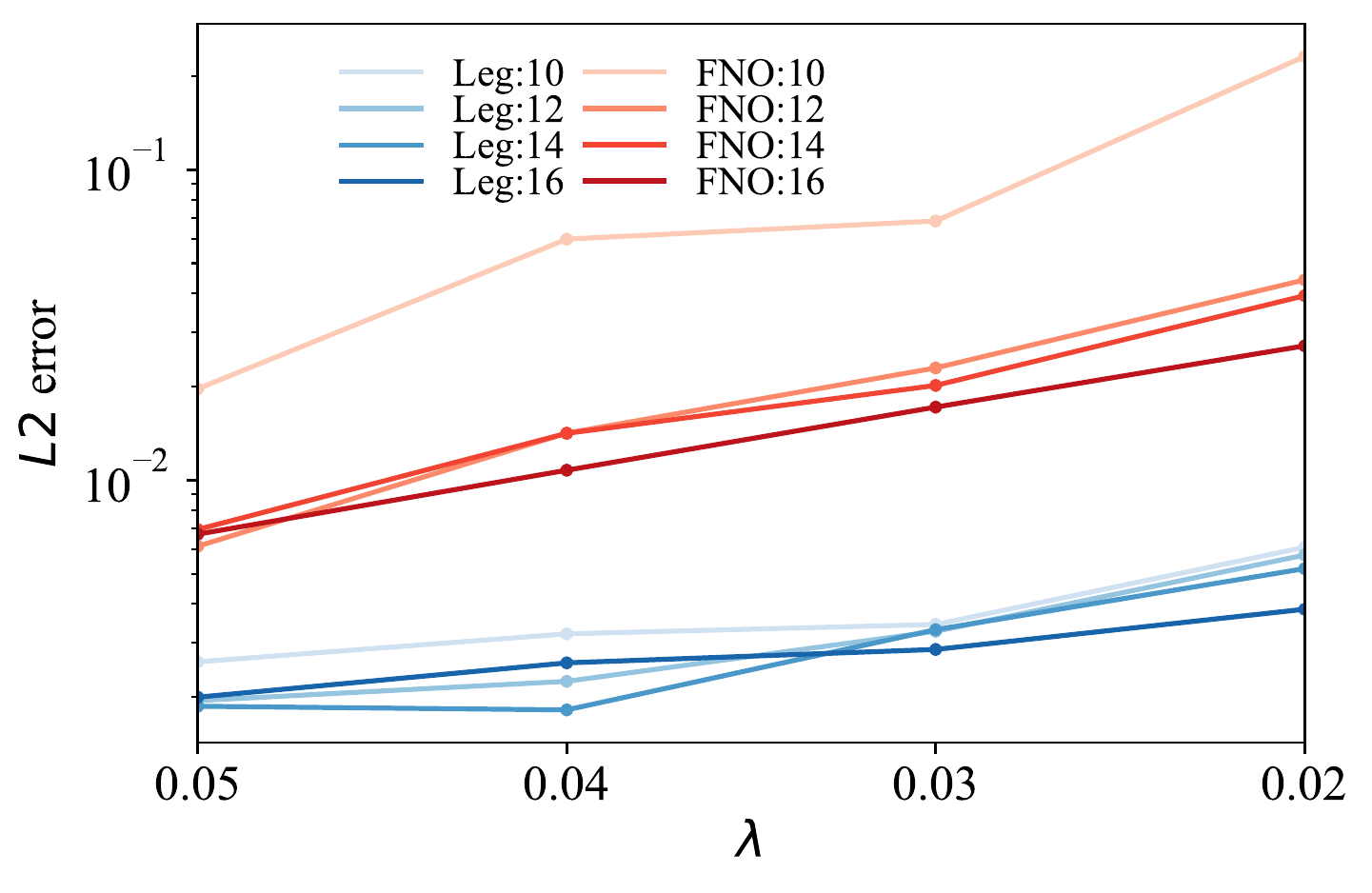}
\caption{Comparing MWT by varying the degree of fluctuations $\lambda$ in the input with resolution $s=1024$. For each convolution, we fix the number of Fourier bases as $k_m$. For FNO, the width is $64$.}
\label{fig:kdv_fluc_acc}
\vspace{-10pt}
\end{wrapfigure}
the relative error is low, which means that a sparse data set with a coarse resolution of $64$ is sufficient for the neural operator to learn the function mapping between infinite-dimensional spaces.

\textbf{Varying fluctuations:} We now vary the
smoothness of the input function $u_{0}(x,0)$ by controlling the parameter $\lambda$, where low values of
$\lambda$ imply more frequent fluctuations and $\lambda\rightarrow 0$ reaches the Brownian motion limit \cite{Filip2019SmoothRandom}. To isolate the importance of incorporating the multiwavelet transformation, we use the same convolution operation as in FNO, i.e., Fourier transform-based convolution with different modes $k_m$ (only single-layer) for $A, B, C$. We see in Figure\,\ref{fig:KdV_fluctuations} that MWT model consistently outperforms the recent baselines for all the values of $\lambda$. A sample input/output from test set is shown in the Figure\,\ref{fig:KdV_fluctuations}. The FNO model with higher values of $k_m$ has better performance due to more Fourier bases for representing the high-frequency signal, while MWT does better even with low modes in its $A, B, C$ CNNs, highlighting the importance of using wavelet-based filters in the signal processing.
\subsection{Theoretical Properties Validation}
\label{ssec:eulerBer}
We test the ability of the proposed MWT model to capture the theoretical properties of the pseudo-differential operator in this Section. Towards that, we consider the Euler-Bernoulli equation \cite{timoshenko1983history} that models the vertical displacement of a finite length beam over time. A Fourier transform version of the beam equation with the constraint of both ends being clamped is as follows
\begin{align}
    \begin{aligned}
        \frac{\partial^{4} u}{\partial x} - \omega^{2}u &= f(x),\quad \frac{\partial u}{\partial x}\Bigr|_{\stackon{$\scriptscriptstyle x=0$}{$\scriptscriptstyle x=1$}} = 0\\
        u(0) &= u(1) = 0,\\
    \end{aligned}
\end{align}
where $u(x)$ is the Fourier transform of the time-varying beam displacement, $\omega$ is the frequency, $f(x)$ is the applied force. The Euler-Bernoulli is a pseudo-differential equation with the maximum derivative order $T+1=4$. We take the task of learning the map from $f$ to $u$. In Figure\,\ref{fig:eulerBernoulli}, we see that for $k\geq3$, the models relative error across epochs is similar, however, they are different for $k<3$, which is in accordance with the Property\,\ref{propr:smoothness}. For $k<3$, the multiwavelets will not be able to annihilate the diagonal of the kernel which is $C^{T-1}$, hence, sparsification cannot occur, and the model learns slow.

\subsection{Burgers' Equation}
\label{ssec:burgers}
The 1-D Burgers' equation is a non-linear PDE occurring in various areas of applied mathematics. For a given field $u(x,t)$ and diffusion coefficient $v$, the 1-D Burgers' equation reads: 
\begin{align}
    \begin{aligned}
        \frac{\partial u}{\partial t}&=-u\frac{\partial u}{\partial x}+v\frac{\partial^2 u}{\partial x^2},  x\in(0,2\pi), t\in(0,1]\\
        u_{0}(x)&=u(x,t=0).
    \end{aligned}
\end{align}
The task for the neural operator is to learn the mapping of initial condition $u(x,t=0)$ to the solutions at $t = 1$ $u(x,t=1)$. To compare with many advanced neural operators under the same conditions, we use the Burgers' data and the results that have been published in ~\cite{li2020fourier} and ~\cite{Li2020MGNO}. The initial condition is sampled as Gaussian random fields where $u_{0}\sim \mathcal N(0,5^4(-\Delta+5^2I)^{-2})$ with periodic boundary conditions. $\Delta$ is the Laplacian, meaning the initial conditions are sampled by sampling its first several coefficients from a Gaussian distribution.
In the Burgers' equation, $v$ is set to $0.1$. The equation is solved with resolution $2^{13}$, and the data with lower resolutions are obtained by sub-sampling the highest resolution data set.

\begin{figure}
\centering
\begin{minipage}{0.49\linewidth}
\centering
\includegraphics[width=\linewidth]{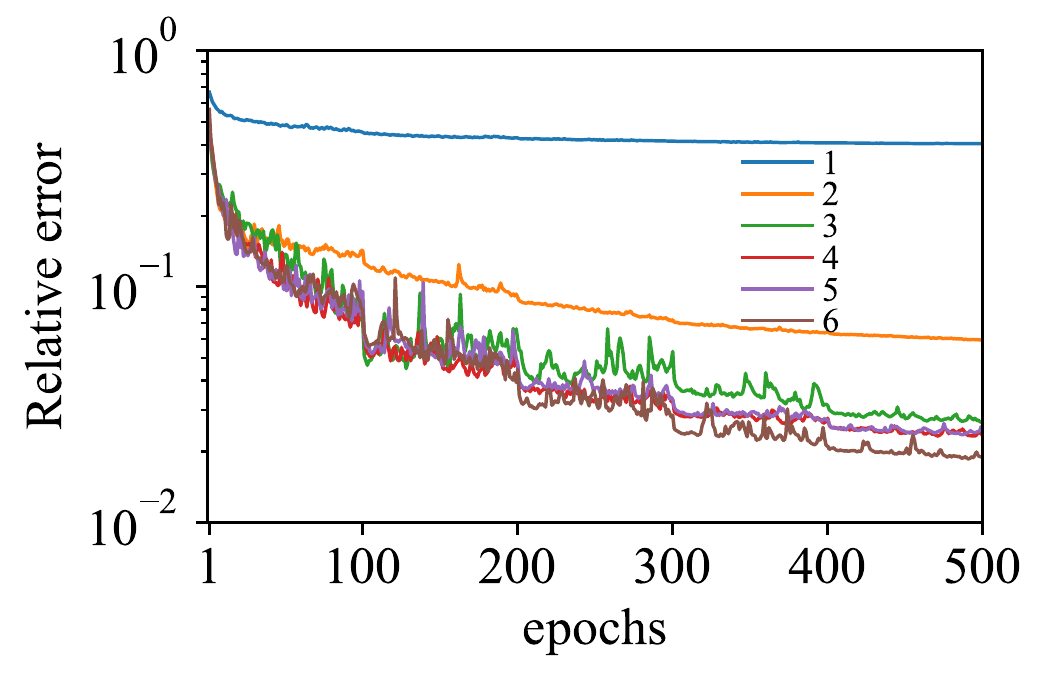}
\caption{Relative $L2$ error vs epochs for MWT Leg with different number of OP basis $k=1,\hdots,6$.}
\label{fig:eulerBernoulli}
\end{minipage}
~
\begin{minipage}{0.49\linewidth}
\centering
\includegraphics[width=\linewidth]{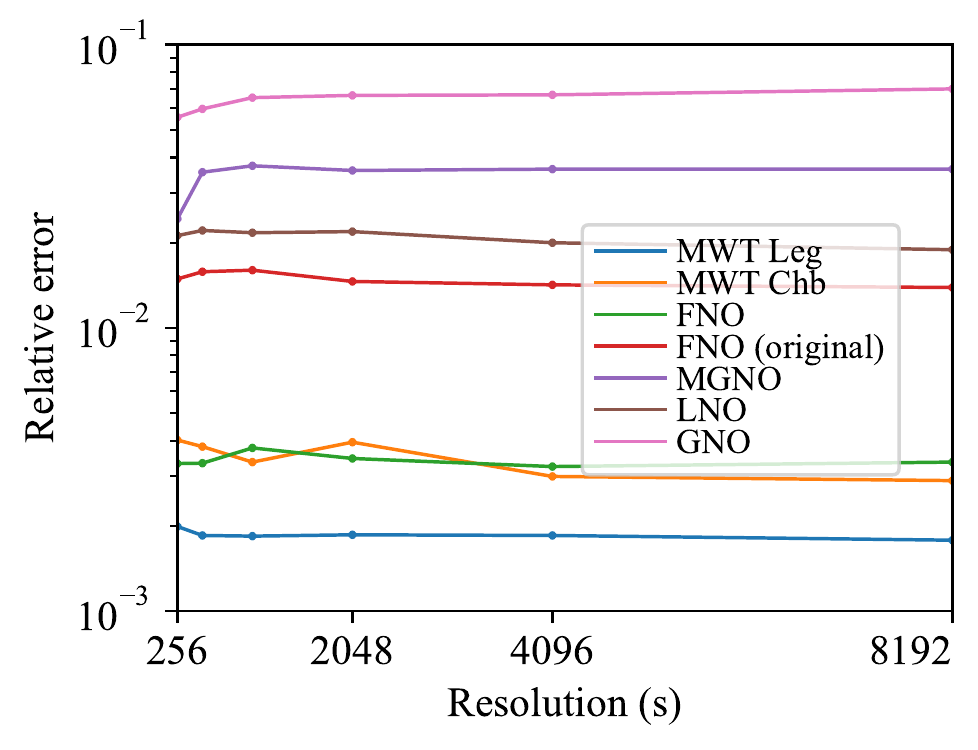}
\caption{Burgers' Equation validation at various input resolution $s$. Our methods: MWT Leg, Chb.}
\label{fig:Burgers}
\end{minipage}
\end{figure}

The results of the experiments on Burgers' equation for different resolutions are shown in Figure\,\ref{fig:Burgers}. Compared to any of the benchmarks, our MWT Leg obtains the lowest relative error, which is an order of magnitude lower than the state-of-the-art. It's worth noting that even in the case of low resolution, MWT Leg still maintains a very low error rate, which shows its potential for learning the function mapping through low-resolution data, that is, the ability to map between infinite-dimensional spaces by learning a limited finite-dimensional spaces mapping.

\subsection{Darcy Flow}
\label{ssec:darcy}
Darcy flow formulated by Darcy\cite{darcy1856fontaines} is one of the basic relationships of hydrogeology, describing the flow of a fluid through a porous medium. We experiment on the steady-state of the 2-d Darcy flow equation on the unit box, where it takes the following form:
\begin{align}
    \begin{aligned}
        \nabla \cdot (a(x)\nabla u(x))&=f(x), &x\in (0,1)^2\\
        u(x)&=0, &x\in \partial(0,1)^2    
    \end{aligned}
\end{align}
We set the experiments to learn the operator mapping the coefficient $a(x)$ to the solution $u(x)$. The coefficients are generated according to $a\sim \mathcal N(0, (-\Delta+3^2I)^{-2})$, where $\Delta$ is the Laplacian with zero Neumann boundary conditions. The threshold of $a(x)$ is set to achieve ellipticity. The solutions $u(x)$ are obtained by using a 2nd-order finite difference scheme on a $512 \times 512$ grid. Data sets of lower resolution are sub-sampled from the original data set.

\begin{table}
\centering
\hspace*{\fill} \\
\hspace*{\fill} \\
\begin{tabular}{p{2cm} p{1.7cm} p{1.7cm} p{1.7cm} p{1.7cm} p{1.7cm} p{1.7cm}}
 \hline
 Networks& s = 32 &s = 64&s = 128&s = 256 & s=512 \\
 \hline
 MWT Leg   &\textbf{0.0152}&\textbf{0.00899}&\textbf{0.00747}&\textbf{0.00722} & \textbf{0.00654} \\
 MWT Chb & 0.0174 & 0.0108 & 0.00872 & 0.00892  & 0.00891\\
 MWT Rnd &0.2435&0.2434&0.2434 &0.2431& 0.2432\\
 \hline
FNO & 0.0177 &  0.0121 &  0.0111 &  0.0107 & 0.0106\\
 MGNO& 0.0501    &0.0519&0.0547&0.0542&-\\
LNO&   0.0524&0.0457&0.0453&0.0428&-\\
 \hline 
\end{tabular}
\caption{\label{tab:darcy}Benchmarks on Darcy Flow equation at various input resolution $s$. Top: Our methods. MWT Rnd instantiate random entries of the filter matrices in \eqref{eqn:scaleDecFilter}-\eqref{eqn:OddScaleRec}. Bottom: prior works on Neural operator.}
\end{table}

The results of the experiments on Darcy Flow for different resolutions are shown in Table\ref{tab:darcy}. MWT Leg again obtains the lowest relative error compared to other neural operators at various resolutions. We also perform an additional experiment, in which the multiwavelet filters $H^{(i)} , G^{(i)}, i=0,1$ are replaced with random values (properly normalized). We see in Table\,\ref{tab:darcy}, that MWT Rnd does not learn the operator map, in fact, its performance is worse than all the models. This signifies the importance of the careful choice of the filter matrices.

\subsection{Additional Experiments}
\label{ssec:addExperi}
Full results for these experiments are provided in the supplementary  materials.

\textbf{Navier Stokes Equation:} The Navier-Stokes (NS) are 2d time-varying PDEs modeling the viscous, incompressible fluids. The proposed MWT model does a $2$d multiwavelet transform for the velocity $u$, while uses a single-layered $3$d convolution for $A, B$ and $C$ to learn dependencies across space-time. We have observed that the proposed MWT Leg is in par with the Sota on the NS equations in Appendix\,\ref{apssec:nsEqn}.

\textbf{Prediction at high resolution:} We show that MWT model trained at lower resolutions for various datasets (for example, training with $s=256$ for Burgers) can predict the output at finer resolutions $s=2048$, with relative error of $0.0226$, thus eliminating the need for expensive sampling. The training and testing with $s=2048$ yields a relative error of $0.00189$. The full experiment is discussed in Appendix\,\ref{apssec:highRes}.

\textbf{Train/evaluation with different sampling rules:} We study the operator learning behavior when the training and evaluation datasets are obtained using random function from different generating rules. In Appendix\,\ref{apssec:diffSample}, the training is done with squared exponential kernel but evaluation is done on different generating rule \cite{Filip2019SmoothRandom} with controllable parameter $\lambda$. 

\section{Conclusion}
\label{sec:concl}
We address the problem of data-driven learning of the operator that maps between two function spaces. Motivated from the fundamental properties of the integral kernel, we found that multiwavelets constitute a natural basis to represent the kernel sparsely. After generalizing the multiwavelets to work with arbitrary measures, we proposed a series of models to learn the integral operator. This work opens up new research directions and possibilities toward designing efficient Neural operators utilizing properties of the kernels, and the suitable basis. We anticipate that the study of this problem will solve many engineering and biological problems such as aircraft wing design, complex fluids dynamics, metamaterials design, cyber-physical systems, neuron-neuron interactions that are modeled by complex PDEs.

\section*{Acknowledgement}
We are thankful to the anonymous reviewers for providing their valuable feedback which improved the manuscript.
We would also like to thank Radu Balan for his valuable feedback. 
We gratefully acknowledge the support by the National Science Foundation Career award under Grant No. CPS/CNS-1453860, the NSF award under Grant CCF-1837131, MCB-1936775, CNS-1932620, the U.S. Army Research Office (ARO) under Grant No. W911NF-17-1-0076, the Okawa Foundation award, and the Defense Advanced Research Projects Agency (DARPA) Young Faculty Award and DARPA Director Award under Grant No. N66001-17-1-4044, an Intel faculty award and a Northrop Grumman grant. A part of this work used the Extreme Science and Engineering Discovery Environment (XSEDE), which is supported by National Science Foundation grant number ACI-1548562. The views, opinions, and/or findings contained in this article are those of the authors and should not be interpreted as representing the official views or policies, either expressed or implied by the Defense Advanced Research Projects Agency, the Army Research Office, the Department of Defense or the National Science Foundation.

\nocite{Gu2020Hippo, li2020fourier, li2020neural, Li2020MGNO, Chou2000Green}

\nocite{Zhu_2018, KHOO_2020, Bhatnagar_2019, bhattacharya2020model, hajij2021algebraicallyinformed, lanthaler2021error, lu2020deeponet, wang2021learning, Lu2021Nature, jagtap2021deep}

\nocite{Belykin1992Wavelets, Belykin1993Fast, BEYLKLN1997Pseudo, Belykin1991FastWavelet, Alpert1991Legendre, Alpert1993L2Bases, ALPERT2002MWT, Daubechies1988, Daubechies1992Lectures}

\nocite{Amartunga2001, Feliu_Fab__2020, fan2019multiscale}

\nocite{Filip2019SmoothRandom, abramowitz1965handbook}

\nocite{meyer1992wavelets, meyer1997wavelets, benedetto1993wavelets, chuong2018pseudodifferential, deng2008harmonic}

\nocite{WANG201466, LI20102276}
\nocite{Chang1983, KHELLAT2006181, HEYDARI2014267, Razzaghi2001}

\nocite{HWANG1982123}

\nocite{WANG20128592, Zhu2017a, Zhu2017b, Tavassoli2009, LI20102284}

\nocite{Kolev_2019, Cotronei1998, kidger2020neural, Chen2020Symplectic, xiong2020neural,rasmussen2006gaussian, boulle2021datadriven}

\nocite{xsede}

\clearpage
\bibliographystyle{plainnat}
\bibliography{references}

\begin{thebibliography}{74}
\providecommand{\natexlab}[1]{#1}
\providecommand{\url}[1]{\texttt{#1}}
\expandafter\ifx\csname urlstyle\endcsname\relax
  \providecommand{\doi}[1]{doi: #1}\else
  \providecommand{\doi}{doi: \begingroup \urlstyle{rm}\Url}\fi

\bibitem[Abramowitz and Stegun(1965)]{abramowitz1965handbook}
M.~Abramowitz and I.A. Stegun.
\newblock \emph{Handbook of Mathematical Functions: With Formulas, Graphs, and
  Mathematical Tables}.
\newblock Applied mathematics series. Dover Publications, 1965.
\newblock ISBN 9780486612720.

\bibitem[Acheson(1991)]{acheson1991elementary}
David~J Acheson.
\newblock Elementary fluid dynamics, 1991.

\bibitem[Adler and Öktem(2017)]{Adler_2017}
Jonas Adler and Ozan Öktem.
\newblock Solving ill-posed inverse problems using iterative deep neural
  networks.
\newblock \emph{Inverse Problems}, 33\penalty0 (12), Nov 2017.
\newblock ISSN 1361-6420.
\newblock \doi{10.1088/1361-6420/aa9581}.
\newblock URL \url{http://dx.doi.org/10.1088/1361-6420/aa9581}.

\bibitem[Alpert et~al.(1993)Alpert, Beylkin, Coifman, and
  Rokhlin]{Belykin1993Fast}
B.~Alpert, G.~Beylkin, R.~Coifman, and V.~Rokhlin.
\newblock Wavelet-like bases for the fast solution of second-kind integral
  equations.
\newblock \emph{SIAM Journal on Scientific Computing}, 14\penalty0
  (1):\penalty0 159--184, 1993.
\newblock \doi{10.1137/0914010}.

\bibitem[Alpert et~al.(2002)Alpert, Beylkin, Gines, and Vozovoi]{ALPERT2002MWT}
B.~Alpert, G.~Beylkin, D.~Gines, and L.~Vozovoi.
\newblock Adaptive solution of partial differential equations in multiwavelet
  bases.
\newblock \emph{Journal of Computational Physics}, 182\penalty0 (1):\penalty0
  149--190, 2002.
\newblock ISSN 0021-9991.

\bibitem[Alpert(1993)]{Alpert1993L2Bases}
Bradley~K. Alpert.
\newblock A class of bases in ${L}^2$ for the sparse representation of integral
  operators.
\newblock \emph{SIAM Journal on Mathematical Analysis}, 24\penalty0
  (1):\penalty0 246--262, 1993.
\newblock \doi{10.1137/0524016}.

\bibitem[Alpert and Rokhlin(1991)]{Alpert1991Legendre}
Bradley~K. Alpert and Vladimir Rokhlin.
\newblock A fast algorithm for the evaluation of legendre expansions.
\newblock \emph{SIAM Journal on Scientific and Statistical Computing},
  12\penalty0 (1):\penalty0 158--179, 1991.
\newblock \doi{10.1137/0912009}.

\bibitem[Amaratunga and Williams(2001)]{Amartunga2001}
Kevin Amaratunga and John Williams.
\newblock Wavelet based green's function approach to 2d pdes.
\newblock \emph{Engineering Computations}, 10, 07 2001.
\newblock \doi{10.1108/eb023913}.

\bibitem[Balan et~al.(2009)Balan, Bodmann, Casazza, and Edidin]{Balan2009}
Radu Balan, Bernhard~G. Bodmann, Peter~G. Casazza, and Dan Edidin.
\newblock Painless reconstruction from magnitudes of frame coefficients.
\newblock \emph{Journal of Fourier Analysis and Applications}, 16:\penalty0
  488–501, 2009.

\bibitem[Batchelor and Batchelor(2000)]{batchelor2000introduction}
Cx~K Batchelor and GK~Batchelor.
\newblock \emph{An introduction to fluid dynamics}.
\newblock Cambridge university press, 2000.

\bibitem[Benedetto(1993)]{benedetto1993wavelets}
J.J. Benedetto.
\newblock \emph{Wavelets: Mathematics and Applications}.
\newblock Studies in Advanced Mathematics. Taylor \& Francis, 1993.
\newblock ISBN 9780849382710.

\bibitem[Beylkin(1992)]{Belykin1992Wavelets}
G.~Beylkin.
\newblock On the representation of operators in bases of compactly supported
  wavelets.
\newblock \emph{SIAM Journal on Numerical Analysis}, 29\penalty0 (6):\penalty0
  1716--1740, 1992.
\newblock \doi{10.1137/0729097}.

\bibitem[Beylkin et~al.(1991)Beylkin, Coifman, and
  Rokhlin]{Belykin1991FastWavelet}
G.~Beylkin, R.~Coifman, and V.~Rokhlin.
\newblock Fast wavelet transforms and numerical algorithms i.
\newblock \emph{Communications on Pure and Applied Mathematics}, 44\penalty0
  (2):\penalty0 141--183, 1991.
\newblock \doi{https://doi.org/10.1002/cpa.3160440202}.

\bibitem[Beylkin and Keiser(1997)]{BEYLKLN1997Pseudo}
Gregory Beylkin and James~M. Keiser.
\newblock An adaptive pseudo-wavelet approach for solving nonlinear partial
  differential equations.
\newblock In \emph{Multiscale Wavelet Methods for Partial Differential
  Equations}, volume~6 of \emph{Wavelet Analysis and Its Applications}, pages
  137--197. Academic Press, 1997.

\bibitem[Bhatnagar et~al.(2019)Bhatnagar, Afshar, Pan, Duraisamy, and
  Kaushik]{Bhatnagar_2019}
Saakaar Bhatnagar, Yaser Afshar, Shaowu Pan, Karthik Duraisamy, and Shailendra
  Kaushik.
\newblock Prediction of aerodynamic flow fields using convolutional neural
  networks.
\newblock \emph{Computational Mechanics}, 64\penalty0 (2):\penalty0 525–545,
  Jun 2019.
\newblock ISSN 1432-0924.
\newblock \doi{10.1007/s00466-019-01740-0}.

\bibitem[Bhattacharya et~al.(2020)Bhattacharya, Hosseini, Kovachki, and
  Stuart]{bhattacharya2020model}
Kaushik Bhattacharya, Bamdad Hosseini, Nikola~B. Kovachki, and Andrew~M.
  Stuart.
\newblock Model reduction and neural networks for parametric pdes, 2020.

\bibitem[Boullé et~al.(2021)Boullé, Earls, and
  Townsend]{boulle2021datadriven}
Nicolas Boullé, Christopher~J. Earls, and Alex Townsend.
\newblock Data-driven discovery of physical laws with human-understandable deep
  learning, 2021.

\bibitem[Boussinesq(1877)]{boussinesq1877essai}
Joseph Boussinesq.
\newblock \emph{Essai sur la th{\'e}orie des eaux courantes}.
\newblock Impr. nationale, 1877.

\bibitem[Chang and Wang(1983)]{Chang1983}
R.~Y. Chang and M.~L. Wang.
\newblock Shifted legendre direct method for variational problems.
\newblock \emph{Journal of Optimization Theory and Applications}, 39\penalty0
  (2):\penalty0 299--307, Feb 1983.
\newblock ISSN 1573-2878.
\newblock \doi{10.1007/BF00934535}.

\bibitem[Chen et~al.(2020)Chen, Zhang, Arjovsky, and
  Bottou]{Chen2020Symplectic}
Zhengdao Chen, Jianyu Zhang, Martin Arjovsky, and Léon Bottou.
\newblock Symplectic recurrent neural networks.
\newblock In \emph{International Conference on Learning Representations}, 2020.
\newblock URL \url{https://openreview.net/forum?id=BkgYPREtPr}.

\bibitem[Chou and Guthart(2000)]{Chou2000Green}
Kenneth~C. Chou and Gary~S. Guthart.
\newblock Representation of green's function integral operators using wavelet
  transforms.
\newblock \emph{Journal of Vibration and Control}, 6\penalty0 (1):\penalty0
  19--48, 2000.
\newblock \doi{10.1177/107754630000600102}.

\bibitem[Chuong(2018)]{chuong2018pseudodifferential}
N.M. Chuong.
\newblock \emph{Pseudodifferential Operators and Wavelets over Real and p-adic
  Fields}.
\newblock Springer International Publishing, 2018.
\newblock ISBN 9783319774732.

\bibitem[Cotronei et~al.(1998)Cotronei, Montefusco, and Puccio]{Cotronei1998}
M.~Cotronei, L.B. Montefusco, and L.~Puccio.
\newblock Multiwavelet analysis and signal processing.
\newblock \emph{IEEE Transactions on Circuits and Systems II: Analog and
  Digital Signal Processing}, 45\penalty0 (8):\penalty0 970--987, 1998.

\bibitem[Darcy(1856)]{darcy1856fontaines}
Henry Darcy.
\newblock \emph{Les fontaines publiques de la ville de Dijon: exposition et
  application...}
\newblock Victor Dalmont, 1856.

\bibitem[Darrigol(2005)]{darrigol2005worlds}
Olivier Darrigol.
\newblock \emph{Worlds of flow: A history of hydrodynamics from the Bernoullis
  to Prandtl}.
\newblock Oxford University Press, 2005.

\bibitem[Daubechies(1988)]{Daubechies1988}
Ingrid Daubechies.
\newblock Orthonormal bases of compactly supported wavelets.
\newblock \emph{Communications on Pure and Applied Mathematics}, 41\penalty0
  (7):\penalty0 909--996, 1988.
\newblock \doi{https://doi.org/10.1002/cpa.3160410705}.

\bibitem[Daubechies(1992)]{Daubechies1992Lectures}
Ingrid Daubechies.
\newblock \emph{Ten Lectures on Wavelets}.
\newblock Society for Industrial and Applied Mathematics, 1992.
\newblock \doi{10.1137/1.9781611970104}.

\bibitem[Deng et~al.(2008)Deng, Meyer, and Han]{deng2008harmonic}
D.~Deng, Y.~Meyer, and Y.~Han.
\newblock \emph{Harmonic Analysis on Spaces of Homogeneous Type}.
\newblock Lecture Notes in Mathematics. Springer Berlin Heidelberg, 2008.
\newblock ISBN 9783540887447.

\bibitem[Driscoll et~al.(2014)Driscoll, Hale, and Trefethen]{Driscoll2014}
T.~A Driscoll, N.~Hale, and L.~N. Trefethen.
\newblock \emph{Chebfun Guide}.
\newblock Pafnuty Publications, 2014.

\bibitem[Fan et~al.(2019)Fan, Feliu-Faba, Lin, Ying, and
  Zepeda-Nunez]{fan2019multiscale}
Yuwei Fan, Jordi Feliu-Faba, Lin Lin, Lexing Ying, and Leonardo Zepeda-Nunez.
\newblock A multiscale neural network based on hierarchical nested bases, 2019.

\bibitem[Feliu-Fabà et~al.(2020)Feliu-Fabà, Fan, and Ying]{Feliu_Fab__2020}
Jordi Feliu-Fabà, Yuwei Fan, and Lexing Ying.
\newblock Meta-learning pseudo-differential operators with deep neural
  networks.
\newblock \emph{Journal of Computational Physics}, 408:\penalty0 109309, May
  2020.
\newblock ISSN 0021-9991.
\newblock \doi{10.1016/j.jcp.2020.109309}.

\bibitem[Filip et~al.(2019)Filip, Javeed, and Trefethen]{Filip2019SmoothRandom}
Silviu-Ioan Filip, Aurya Javeed, and Lloyd Trefethen.
\newblock Smooth random functions, random odes, and gaussian processes.
\newblock \emph{SIAM Review}, 61:\penalty0 185--205, 01 2019.
\newblock \doi{10.1137/17M1161853}.

\bibitem[Greenfeld et~al.(2019)Greenfeld, Galun, Kimmel, Yavneh, and
  Basri]{greenfeld2019learning}
Daniel Greenfeld, Meirav Galun, Ron Kimmel, Irad Yavneh, and Ronen Basri.
\newblock Learning to optimize multigrid pde solvers, 2019.

\bibitem[Gu et~al.(2020)Gu, Dao, Ermon, Rudra, and R\'{e}]{Gu2020Hippo}
Albert Gu, Tri Dao, Stefano Ermon, Atri Rudra, and Christopher R\'{e}.
\newblock Hippo: Recurrent memory with optimal polynomial projections.
\newblock In \emph{Advances in Neural Information Processing Systems},
  volume~33, pages 1474--1487, 2020.

\bibitem[Guo et~al.(2016)Guo, Li, and Iorio]{Guo2016CNN}
Xiaoxiao Guo, Wei Li, and Francesco Iorio.
\newblock Convolutional neural networks for steady flow approximation.
\newblock In \emph{Proceedings of the 22nd ACM SIGKDD International Conference
  on Knowledge Discovery and Data Mining}, KDD '16, page 481–490. Association
  for Computing Machinery, 2016.

\bibitem[Hajij et~al.(2021)Hajij, Zamzmi, Dawson, and
  Muller]{hajij2021algebraicallyinformed}
Mustafa Hajij, Ghada Zamzmi, Matthew Dawson, and Greg Muller.
\newblock Algebraically-informed deep networks (aidn): A deep learning approach
  to represent algebraic structures, 2021.

\bibitem[Heydari et~al.(2014)Heydari, Hooshmandasl, and
  Mohammadi]{HEYDARI2014267}
M.H. Heydari, M.R. Hooshmandasl, and F.~Mohammadi.
\newblock Legendre wavelets method for solving fractional partial differential
  equations with dirichlet boundary conditions.
\newblock \emph{Applied Mathematics and Computation}, 234:\penalty0 267--276,
  2014.
\newblock ISSN 0096-3003.
\newblock \doi{https://doi.org/10.1016/j.amc.2014.02.047}.

\bibitem[Hwang and Shih(1982)]{HWANG1982123}
Chyi Hwang and Yen-Ping Shih.
\newblock Solution of integral equations via laguerre polynomials.
\newblock \emph{Computers \& Electrical Engineering}, 9\penalty0 (3):\penalty0
  123--129, 1982.
\newblock ISSN 0045-7906.
\newblock \doi{https://doi.org/10.1016/0045-7906(82)90018-0}.

\bibitem[Jagtap et~al.(2021)Jagtap, Shin, Kawaguchi, and
  Karniadakis]{jagtap2021deep}
Ameya~D. Jagtap, Yeonjong Shin, Kenji Kawaguchi, and George~Em Karniadakis.
\newblock Deep kronecker neural networks: A general framework for neural
  networks with adaptive activation functions, 2021.

\bibitem[Kajani et~al.(2009)Kajani, Vencheh, and Ghasemi]{Tavassoli2009}
M.~Tavassoli Kajani, A.~Hadi Vencheh, and M.~Ghasemi.
\newblock The chebyshev wavelets operational matrix of integration and product
  operation matrix.
\newblock \emph{International Journal of Computer Mathematics}, 86\penalty0
  (7):\penalty0 1118--1125, 2009.
\newblock \doi{10.1080/00207160701736236}.

\bibitem[Khellat and Yousefi(2006)]{KHELLAT2006181}
F.~Khellat and S.A. Yousefi.
\newblock The linear legendre mother wavelets operational matrix of integration
  and its application.
\newblock \emph{Journal of the Franklin Institute}, 343\penalty0 (2):\penalty0
  181--190, 2006.
\newblock ISSN 0016-0032.
\newblock \doi{https://doi.org/10.1016/j.jfranklin.2005.11.002}.

\bibitem[Khoo et~al.(2020)Khoo, Lu, and Ying]{KHOO_2020}
Yuehaw Khoo, Jianfeng Lu, and Lexing Ying.
\newblock Solving parametric pde problems with artificial neural networks.
\newblock \emph{European Journal of Applied Mathematics}, 32\penalty0
  (3):\penalty0 421–435, Jul 2020.
\newblock ISSN 1469-4425.

\bibitem[Kidger et~al.(2020)Kidger, Morrill, Foster, and
  Lyons]{kidger2020neural}
Patrick Kidger, James Morrill, James Foster, and Terry Lyons.
\newblock Neural controlled differential equations for irregular time series,
  2020.

\bibitem[Kochkov et~al.(2021)Kochkov, Smith, Alieva, Wang, Brenner, and
  Hoyer]{Kochkove2101784118}
Dmitrii Kochkov, Jamie~A. Smith, Ayya Alieva, Qing Wang, Michael~P. Brenner,
  and Stephan Hoyer.
\newblock Machine learning{\textendash}accelerated computational fluid
  dynamics.
\newblock \emph{Proceedings of the National Academy of Sciences}, 118\penalty0
  (21), 2021.
\newblock ISSN 0027-8424.
\newblock \doi{10.1073/pnas.2101784118}.

\bibitem[Kolev et~al.(2019)Kolev, Cooklev, and Keinert]{Kolev_2019}
Vasil Kolev, Todor Cooklev, and Fritz Keinert.
\newblock Design of a simple orthogonal multiwavelet filter by matrix spectral
  factorization.
\newblock \emph{Circuits, Systems, and Signal Processing}, 39\penalty0
  (4):\penalty0 2006–2041, Aug 2019.

\bibitem[Lanthaler et~al.(2021)Lanthaler, Mishra, and
  Karniadakis]{lanthaler2021error}
Samuel Lanthaler, Siddhartha Mishra, and George~Em Karniadakis.
\newblock Error estimates for deeponets: A deep learning framework in infinite
  dimensions, 2021.

\bibitem[LI(2010)]{LI20102284}
Yuanlu LI.
\newblock Solving a nonlinear fractional differential equation using chebyshev
  wavelets.
\newblock \emph{Communications in Nonlinear Science and Numerical Simulation},
  15\penalty0 (9):\penalty0 2284--2292, 2010.
\newblock ISSN 1007-5704.
\newblock \doi{https://doi.org/10.1016/j.cnsns.2009.09.020}.

\bibitem[Li and Zhao(2010)]{LI20102276}
Yuanlu Li and Weiwei Zhao.
\newblock Haar wavelet operational matrix of fractional order integration and
  its applications in solving the fractional order differential equations.
\newblock \emph{Applied Mathematics and Computation}, 216\penalty0
  (8):\penalty0 2276--2285, 2010.
\newblock ISSN 0096-3003.
\newblock \doi{https://doi.org/10.1016/j.amc.2010.03.063}.

\bibitem[Li et~al.(2020{\natexlab{a}})Li, Kovachki, Azizzadenesheli, Liu,
  Bhattacharya, Stuart, and Anandkumar]{li2020fourier}
Zongyi Li, Nikola Kovachki, Kamyar Azizzadenesheli, Burigede Liu, Kaushik
  Bhattacharya, Andrew Stuart, and Anima Anandkumar.
\newblock Fourier neural operator for parametric partial differential
  equations, 2020{\natexlab{a}}.

\bibitem[Li et~al.(2020{\natexlab{b}})Li, Kovachki, Azizzadenesheli, Liu,
  Bhattacharya, Stuart, and Anandkumar]{li2020neural}
Zongyi Li, Nikola Kovachki, Kamyar Azizzadenesheli, Burigede Liu, Kaushik
  Bhattacharya, Andrew Stuart, and Anima Anandkumar.
\newblock Neural operator: Graph kernel network for partial differential
  equations, 2020{\natexlab{b}}.

\bibitem[Li et~al.(2020{\natexlab{c}})Li, Kovachki, Azizzadenesheli, Liu,
  Stuart, Bhattacharya, and Anandkumar]{Li2020MGNO}
Zongyi Li, Nikola Kovachki, Kamyar Azizzadenesheli, Burigede Liu, Andrew
  Stuart, Kaushik Bhattacharya, and Anima Anandkumar.
\newblock Multipole graph neural operator for parametric partial differential
  equations.
\newblock In \emph{Advances in Neural Information Processing Systems},
  volume~33, pages 6755--6766, 2020{\natexlab{c}}.

\bibitem[Lu et~al.(2020)Lu, Jin, and Karniadakis]{lu2020deeponet}
Lu~Lu, Pengzhan Jin, and George~Em Karniadakis.
\newblock Deeponet: Learning nonlinear operators for identifying differential
  equations based on the universal approximation theorem of operators, 2020.

\bibitem[Lu et~al.(2021)Lu, Jin, Pang, Zhang, and Karniadakis]{Lu2021Nature}
Lu~Lu, Pengzhan Jin, Guofei Pang, Zhongqiang Zhang, and George~Em Karniadakis.
\newblock Learning nonlinear operators via deeponet based on the universal
  approximation theorem of operators.
\newblock \emph{Nature Machine Intelligence}, 3\penalty0 (3):\penalty0
  218--229, Mar 2021.
\newblock ISSN 2522-5839.

\bibitem[McKeown et~al.(2020)McKeown, Ostilla-M{\'o}nico, Pumir, Brenner, and
  Rubinstein]{McKeowneaaz2717}
Ryan McKeown, Rodolfo Ostilla-M{\'o}nico, Alain Pumir, Michael~P. Brenner, and
  Shmuel~M. Rubinstein.
\newblock Turbulence generation through an iterative cascade of the elliptical
  instability.
\newblock \emph{Science Advances}, 6\penalty0 (9), 2020.
\newblock \doi{10.1126/sciadv.aaz2717}.

\bibitem[Meyer and Salinger(1992)]{meyer1992wavelets}
Y.~Meyer and D.H. Salinger.
\newblock \emph{Wavelets and Operators: Volume 1}.
\newblock Cambridge Studies in Advanced Mathematics. Cambridge University
  Press, 1992.
\newblock ISBN 9780521458696.

\bibitem[Meyer et~al.(1997)Meyer, Coifman, and Salinger]{meyer1997wavelets}
Y.~Meyer, R.~Coifman, and D.~Salinger.
\newblock \emph{Wavelets: Calder{\'o}n-Zygmund and Multilinear Operators}.
\newblock Cambridge Studies in Advanced Mathematics. Cambridge University
  Press, 1997.
\newblock ISBN 9780521420013.

\bibitem[Ovarlez et~al.(2020)Ovarlez, Vu~Nguyen~Le, Smit, Fall, Mari,
  Chatt{\'e}, and Colin]{Ovarlezeaay5589}
Guillaume Ovarlez, Anh Vu~Nguyen~Le, Wilbert~J. Smit, Abdoulaye Fall, Romain
  Mari, Guillaume Chatt{\'e}, and Annie Colin.
\newblock Density waves in shear-thickening suspensions.
\newblock \emph{Science Advances}, 6\penalty0 (16), 2020.
\newblock \doi{10.1126/sciadv.aay5589}.

\bibitem[Patel et~al.(2021)Patel, Trask, Wood, and Cyr]{PATEL2021113500}
Ravi~G. Patel, Nathaniel~A. Trask, Mitchell~A. Wood, and Eric~C. Cyr.
\newblock A physics-informed operator regression framework for extracting
  data-driven continuum models.
\newblock \emph{Computer Methods in Applied Mechanics and Engineering},
  373:\penalty0 113500, 2021.
\newblock ISSN 0045-7825.
\newblock \doi{https://doi.org/10.1016/j.cma.2020.113500}.

\bibitem[Raissi et~al.(2019)Raissi, Perdikaris, and Karniadakis]{RAISSI2019686}
M.~Raissi, P.~Perdikaris, and G.E. Karniadakis.
\newblock Physics-informed neural networks: A deep learning framework for
  solving forward and inverse problems involving nonlinear partial differential
  equations.
\newblock \emph{Journal of Computational Physics}, 378:\penalty0 686--707,
  2019.
\newblock ISSN 0021-9991.

\bibitem[Rasmussen and Williams(2006)]{rasmussen2006gaussian}
C.E. Rasmussen and C.K.I. Williams.
\newblock \emph{Gaussian Processes for Machine Learning}.
\newblock Adaptative computation and machine learning series. University Press
  Group Limited, 2006.

\bibitem[Razzaghi and Yousefi(2001)]{Razzaghi2001}
Mohsen Razzaghi and Samira Yousefi.
\newblock The legendre wavelets operational matrix of integration.
\newblock \emph{International Journal of Systems Science}, 32:\penalty0
  495--502, 04 2001.
\newblock \doi{10.1080/00207720120227}.

\bibitem[Shlesinger et~al.(1987)Shlesinger, West, and Klafter]{Shlesinger1987}
M.~F. Shlesinger, B.~J. West, and J.~Klafter.
\newblock L\'evy dynamics of enhanced diffusion: Application to turbulence.
\newblock \emph{Phys. Rev. Lett.}, 58:\penalty0 1100--1103, Mar 1987.
\newblock \doi{10.1103/PhysRevLett.58.1100}.

\bibitem[Silverman et~al.(1999)Silverman, Vassilicos, and
  Kingsbury]{Silverman1999}
B.~W. Silverman, J.~C. Vassilicos, and Nick Kingsbury.
\newblock Image processing with complex wavelets.
\newblock \emph{Philosophical Transactions of the Royal Society of London.
  Series A: Mathematical, Physical and Engineering Sciences}, 357\penalty0
  (1760):\penalty0 2543--2560, 1999.
\newblock \doi{10.1098/rsta.1999.0447}.

\bibitem[Stoer et~al.(2002)Stoer, Bartels, Gautschi, Bulirsch, and
  Witzgall]{stoer2002introduction}
J.~Stoer, R.~Bartels, W.~Gautschi, R.~Bulirsch, and C.~Witzgall.
\newblock \emph{Introduction to Numerical Analysis}.
\newblock Texts in Applied Mathematics. Springer New York, 2002.
\newblock ISBN 9780387954523.

\bibitem[Timoshenko(1983)]{timoshenko1983history}
S.~Timoshenko.
\newblock \emph{History of Strength of Materials: With a Brief Account of the
  History of Theory of Elasticity and Theory of Structures}.
\newblock Dover Civil and Mechanical Engineering Series. Dover Publications,
  1983.
\newblock ISBN 9780486611877.

\bibitem[Towns et~al.(2014)Towns, Cockerill, Dahan, Foster, Gaither, Grimshaw,
  Hazlewood, Lathrop, Lifka, Peterson, Roskies, Scott, and
  Wilkins-Diehr]{xsede}
J.~Towns, T.~Cockerill, M.~Dahan, I.~Foster, K.~Gaither, A.~Grimshaw,
  V.~Hazlewood, S.~Lathrop, D.~Lifka, G.~D. Peterson, R.~Roskies, J.~R. Scott,
  and N.~Wilkins-Diehr.
\newblock Xsede: Accelerating scientific discovery.
\newblock \emph{Computing in Science \& Engineering}, 16\penalty0 (5):\penalty0
  62--74, Sept.-Oct. 2014.
\newblock ISSN 1521-9615.
\newblock \doi{10.1109/MCSE.2014.80}.
\newblock URL \url{doi.ieeecomputersociety.org/10.1109/MCSE.2014.80}.

\bibitem[Wang et~al.(2014)Wang, Ma, and Meng]{WANG201466}
Lifeng Wang, Yunpeng Ma, and Zhijun Meng.
\newblock Haar wavelet method for solving fractional partial differential
  equations numerically.
\newblock \emph{Applied Mathematics and Computation}, 227:\penalty0 66--76,
  2014.
\newblock ISSN 0096-3003.
\newblock \doi{https://doi.org/10.1016/j.amc.2013.11.004}.

\bibitem[Wang et~al.(2021)Wang, Wang, and Perdikaris]{wang2021learning}
Sifan Wang, Hanwen Wang, and Paris Perdikaris.
\newblock Learning the solution operator of parametric partial differential
  equations with physics-informed deeponets, 2021.

\bibitem[Wang and Fan(2012)]{WANG20128592}
Yanxin Wang and Qibin Fan.
\newblock The second kind chebyshev wavelet method for solving fractional
  differential equations.
\newblock \emph{Applied Mathematics and Computation}, 218\penalty0
  (17):\penalty0 8592--8601, 2012.
\newblock ISSN 0096-3003.
\newblock \doi{https://doi.org/10.1016/j.amc.2012.02.022}.

\bibitem[Xiong et~al.(2020)Xiong, He, Tong, and Zhu]{xiong2020neural}
Shiying Xiong, Xingzhe He, Yunjin Tong, and Bo~Zhu.
\newblock Neural vortex method: from finite lagrangian particles to infinite
  dimensional eulerian dynamics, 2020.

\bibitem[Xue and Bogdan(2017)]{YuankunICCPS2016}
Yuankun Xue and Paul Bogdan.
\newblock Constructing compact causal mathematical models for complex dynamics.
\newblock In \emph{2017 ACM/IEEE 8th International Conference on Cyber-Physical
  Systems (ICCPS)}, pages 97--108, April 2017.

\bibitem[Zhu and Wang(2017{\natexlab{a}})]{Zhu2017a}
Li~Zhu and Yanxin Wang.
\newblock Solving fractional partial differential equations by using the second
  chebyshev wavelet operational matrix method.
\newblock \emph{Nonlinear Dynamics}, 89\penalty0 (3):\penalty0 1915--1925, Aug
  2017{\natexlab{a}}.
\newblock ISSN 1573-269X.
\newblock \doi{10.1007/s11071-017-3561-7}.

\bibitem[Zhu and Wang(2017{\natexlab{b}})]{Zhu2017b}
Li~Zhu and Yanxin Wang.
\newblock Solving fractional partial differential equations by using the second
  chebyshev wavelet operational matrix method.
\newblock \emph{Nonlinear Dynamics}, 89\penalty0 (3):\penalty0 1915--1925, Aug
  2017{\natexlab{b}}.
\newblock ISSN 1573-269X.
\newblock \doi{10.1007/s11071-017-3561-7}.

\bibitem[Zhu and Zabaras(2018)]{Zhu_2018}
Yinhao Zhu and Nicholas Zabaras.
\newblock Bayesian deep convolutional encoder–decoder networks for surrogate
  modeling and uncertainty quantification.
\newblock \emph{Journal of Computational Physics}, 366:\penalty0 415–447, Aug
  2018.
\newblock ISSN 0021-9991.

\end{thebibliography}

\clearpage

\appendix
\section{Technical Background}
\label{apsec:techBack}
We present here some technical preliminaries that are used in the current work. The literature for some of the topics is vast, and we list only the properties that are useful specifically for this work.

\subsection{Wavelets}
\label{apssec:wavelets}
The wavelets represent sets of functions that result from dilation and translation from a single function, often termed as `mother function', or `mother wavelet'. For a given mother wavelet $\psi(x)$, the resulting wavelets are written as 
\begin{equation}
    \psi_{a,b}(x) = \frac{1}{\vert a\vert^{1/2}}\psi\left(\frac{x-b}{a} \right), \qquad a, b\in \mathbb{R}, a\neq 0, x\in D,
\end{equation}
where $a, b$ are the dilation, translation factor, respectively, and $D$ is the domain of the wavelets under consideration. In this work, we are interested in the compactly supported wavelets, or $D$ is a finite interval $[l, r]$, and we also take $\psi\in L^2$. The consideration for non-compact wavelets, for example, Daubechies \cite{Daubechies1988} will be a future consideration. For the rest of this work, without loss of generality, we restrict ourself to the finite domain $D = [0, 1]$, and extension to any $[l, r]$ can be simply done by making suitable shift and scale. 

\begin{figure*}
	\centering
	\includegraphics[width=\linewidth]{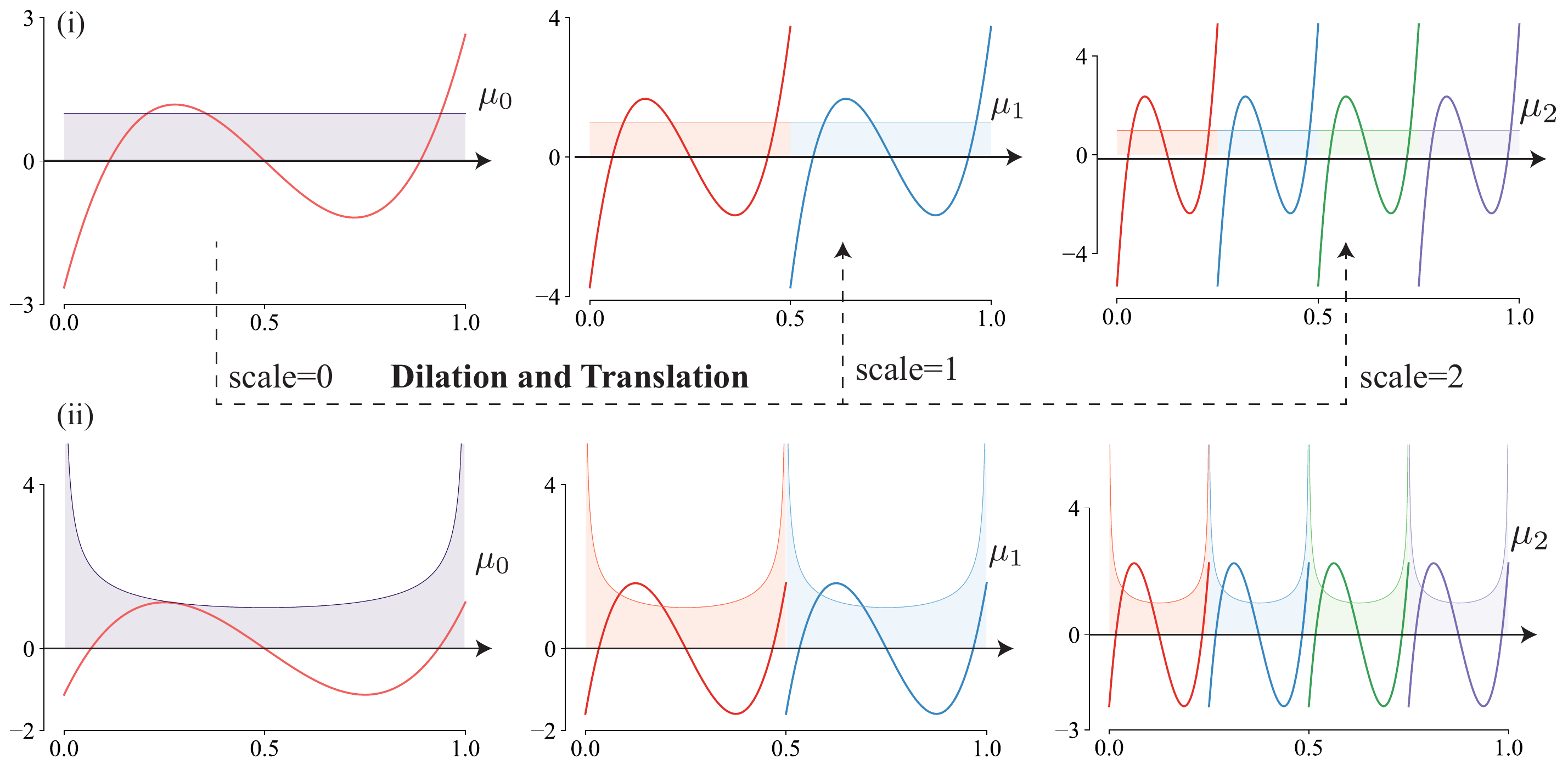}
	\caption{\textbf{Wavelet dilation and translation}. The dilation and translation of the mother wavelet function from left to right. The scale$=0$ represents the mother wavelet function with its measure $\mu_{0}$. The higher scales $(1, 2)$ are obtained by scale/shift with a factor of $2$. \textbf{(i)} Mother wavelet using shifted Legendre polynomial $P_{3}(2x-1)$ with the uniform measure $\mu_{0}$, while \textbf{(ii)} uses shifted Chebyshev polynomial $T_{3}(2x-1)$ with the non-uniform measure $\mu_{0}$.}
	\label{apfig:wavelDemo}
\end{figure*}

From a numerical perspective, discrete values (or Discrete Wavelet Transform) of $a, b$ are more useful, and hence, we take $a = 2^{-j}, j = 0, 1, \hdots, L-1$, where $L$ are the finite number of scales up to which the dilation occurs, and the dilation factor is $2$. For a given value of $a = 2^{-j}$, the values of $b$ can be chosen as , $b = na,\quad n = 0, 1, \hdots, 2^j-1$. The resulting wavelets are now expressed as $\psi_{j,n}(x) = 2^{j/2}\psi(2^j x-n),\quad n = 0, 1, \hdots 2^j-1$, and $x\in [n2^{-j}, (n+1)2^{-j}]$. Given a mother wavelet function, the dilation and translation operations for three scales $(L=3)$ is shown in Figure\,\ref{apfig:wavelDemo}. For a given function $f$, the discrete wavelet transform is obtained by projecting the function $f$ onto the wavelets $\psi_{j,n}$ as
\begin{equation}
    c_{j,n} = \int\nolimits_{n2^{-j}}^{(n+1)2^{-j}}f(x)\psi_{j,n}dx,
\end{equation}
where $c_{j,n}$ are the discrete wavelet transform coefficients.

\subsection{Orthogonal Polynomials}
\label{apssec:orthPoly}
The next set of ingredients that are useful to us are the family of orthogonal polynomials (OPs). Specifically, the OPs in the current work will serve as the mother wavelets or span the 'mother subspace' (see Section\,\ref{apssec:multiwave}). Therefore, we are interested in the OPs that are non-zero over a finite domain, and are zero almost everywhere (a.e.). For a given measure $\mu$ that defines the OPs, a sequence of OPs $P_{0}(x), P_{1}(x), \hdots$ satisfy deg$(P_{i}) = i$, and $\langle P_{i}, P_{j}\rangle_{\mu} = 0, \forall i\neq j$, where $\langle P_{i}, P_{j}\rangle_{\mu} = \int P_{i}(x)P_{j}(x)d\mu$. Therefore, sequence of OPs are particularly useful as they can act as a set of basis for the space of polynomials with degree < $d$ by using $P_{0}, \hdots, P_{d-1}(x)$.

The popular set of OPs are hypergeometric polynomials (also known as Jacobi polynomials). Among them, the common choices are Legendre, Chebyshev, and Gegenbauer (which generalize Legendre and Chebyshev) polynomials. These polynomials are defined on a finite interval of $[-1, 1]$ and are useful for the current work. The other set of OPs are Laguerre, and Hermite polynomials which are defined over non-finite domain. Such OPs could be used for extending the current work to non-compact wavelets. We now review some defining properties of the Legendre and Chebyshev polynomials.

\subsubsection{Legendre Polynomials}
\label{apsssec:legendre}
The Legendre polynomials are defined with respect to (w.r.t.) a uniform weight function $w_{L}(x) = 1$ for $-1\leq x\leq 1$ or $w_{L}(x) = {\bf 1}_{[-1, 1]}(x)$ such that
\begin{equation}
    \int\limits_{-1}^{1}P_{i}(x)P_{j}(x)dx = 
    \begin{cases}
    \frac{2}{2i+1} & i = j,\\
    0 & i\neq j.
    \end{cases}
\end{equation}
For our purpose, we shift and scale the Legendre polynomials such that they are defined over $[0, 1]$ as $P_{i}(2x-1)$, and the corresponding weight function as $w_{L}(2x-1)$. 

\textbf{Derivatives:} The Legendre polynomials satisfy the following recurrence relationships
\begin{align*}
    iP_{i}(x) &= (2i-1)xP_{i-1}(x) - (i-1)P_{i-2}(x),\\
    (2i+1)P_{i}(x) &= P_{i+1}^{\prime}(x) - P_{i-1}^{\prime}(x),
\end{align*}
which allow us to express the derivatives as a linear combination of lower-degree polynomials itself as follows:
\begin{equation}
    P_{i}^{\prime}(x) = (2i-1)P_{i-1}(x) + (2i-3)P_{i-1}(x) + \hdots,
\end{equation}
where the summation ends at either $P_{0}(x)$ or $P_{1}(x)$, with $P_{0}(x) = 1$ and $P_{1}(x) = x$. 

\textbf{Basis:} A set of orthonormal basis of the space of polynomials with degree $<d$ defined over the interval $[0, 1]$ is obtained using shifted Legendre polynomials such that
\begin{align*}
    \phi_{i} = \sqrt{2i+1}P_{i}(2x-1),
\end{align*}
w.r.t. weight function $w(x) = w_{L}(2x-1)$, such that
\begin{equation*}
    \langle \phi_{i}, \phi_{j}\rangle_{\mu} = \int\nolimits_{0}^{1}\phi_{i}(x)\phi_{j}(x)dx = \delta_{ij}.
\end{equation*}

\subsubsection{Chebyshev Polynomials}
\label{apsssec:chebyshev}

The Chebyshev polynomials are two sets of polynomial sequences (first, second order) as $T_{i}, U_{i}$. We take the polynomial of the first order $T_{i}(x)$ of degree $i$ which is defined w.r.t. weight function $w_{Ch}(x) = 1/\sqrt{1-x^2}$ for $-1\leq x\leq 1$ as
\begin{equation}
\int\limits_{-1}^{1}T_{i}(x)T_{j}(x)\frac{1}{\sqrt{1-x^2}}dx = \begin{cases}
\pi & i = j = 0, \\
\pi/2 & i = j > 0,\\
0 & i\neq j.
\end{cases}
\end{equation}
After applying the scale and shift to the Chebyshev polynomials such that their domain is limited to $[0 ,1]$, we get $T_{i}(2x-1)$ and the associated weight function as $w_{Ch}(2x-1)$ such that $T_{i}(2x-1)$ are orthogonal w.r.t. $w_{Ch}(2x-1)$ over the interval $[0, 1]$.

\textbf{Derivatives:} The Chebyshev polynomials of the first order satisfy the following recurrence relationships
\begin{align*}
    2T_{i}(x) &= \frac{1}{i+1}T_{i+1}^{\prime}(x) - \frac{1}{i-1}T_{i-1}^{\prime}(x), \quad i>1,\\
    T_{i+1}(x) &= 2xT_{i}(x) - T_{i-1}(x),\\
\end{align*}
The derivative of the $T_{i}(x)$ can be written as the following summation of sequence of lower degree polynomials
\begin{align*}
    T_{i}^{\prime}(x) &= i(2T_{i-1}(x) + 2T_{i-3}(x) + \hdots),
\end{align*}
where the series ends at either $T_{0}(x) = 1$, or $T_{1}(x) = x$. Alternatively, the derivative of $T_{i}(x)$ can also be written as $T_{i}^{\prime}(x) = iU_{i-1}(x)$, where $U_{i}(x)$ is the second-order Chebyshev polynomial of degree $i$.

\textbf{Basis:} A set of orthonormal basis of the space of polynomials of degree up to $d$ and domain $[0, 1]$ is obtained using Chebyshev polynomials as
\begin{align*}
    \phi_{i} = 
    \begin{cases}
    \frac{2}{\sqrt{\pi}}T_{i}(2x-1) & i>0,\\
    \sqrt{\frac{2}{\pi}} & i = 0.
    \end{cases}
\end{align*}
w.r.t. weight function $w_{Ch}(2x-1)$, or 
\begin{equation*}
    \langle \phi_{i}, \phi_{j}\rangle_{\mu} = \int\nolimits_{0}^{1}\phi_{i}(x)\phi_{j}(x)w_{Cb}(2x-1)dx = \delta_{ij}.
\end{equation*}

\textbf{Roots:} Another useful property of Chebyshev polynomials is that they can be expressed as trigonometric functions; specifically, $T_{n}(\cos\theta) = \cos(n\theta)$. The roots of such are also well-defined in the interval $[-1, 1]$. For $T_{n}(x)$, the $n$ roots $x_{1}, \hdots, x_{n}$ are given by
\begin{equation*}
    x_{i} = \cos\left(\frac{\pi}{n}\left(i - \frac{1}{2}\right)\right).
\end{equation*}

\subsection{Multiwavelets}
\label{apssec:multiwave}
The multiwavelets, as introduced in \cite{Alpert1993L2Bases}, exploit the advantages of both wavelets (Section\,\ref{apssec:wavelets}) as well as OPs (Section\,\ref{apssec:orthPoly}). For a given function $f$, instead of projecting the function onto a single wavelet function (wavelet transform), the multiwavelets go one step further and projects the function onto a subspace of degree-restricted polynomials. Along the essence of the wavelet-transform, in multiwavelets, a sequence of wavelet bases are constructed which are scaled/shifted version of the basis of the coarsest scale polynomial subspace. 

In this work, we present a measure-version of the multiwavelets which opens-up a family of the multiwavelet-based models for the operator learning. In Section\,~\ref{apssec:filtersProj}, we provide a detailed mathematical formulation for developing multiwavelets using any set of OPs with measures which can be non-uniform. To be able to develop compactly supported multiwavelets, we have restricted ourself to the family of OPs which are non-zero only over a finite interval. The extension to non-compact wavelets could be done by using OPs which are non-zero over complete/semi range of the real-axis (for example, Laguerre, Hermite polynomials). As an example, we present the expressions for Legendre polynomials which use uniform measure in Section\,\ref{apssec:legendre}, and Chebyshev polynomials which use non-uniform measure in Section\,\ref{apssec:chebyshev}. The work can be readily extended to other family of OPs like Gegenbauer polynomials.

\subsection{Pseudo-Differential Equations}
\label{apssec:pseudoDiffOper}
The linear inhomogeneous pseudo-differential equations $\mathcal{L}u = f$ have the operator which takes the following form
\begin{equation}
    \mathcal{L} = \sum\nolimits_{\alpha\in\mathbb{A}}a_{\alpha}(x)\partial^{\alpha}_{x},
    \label{apeqn:pseudoDiffCanonical}
\end{equation}
where $\mathbb{A}$ is the subset of natural numbers $\mathbb{N}\cup\{0\}$, and $x\in\mathbb{R}^{n}$. The order of the equation is denoted by the highest integer in the set $\mathbb{A}$. The simplest and the most useful case of pseudo-differential operators $\mathcal{L}$ is the one in which $a_{\alpha}(x)\in C^{\infty}$. In the pseudo-differential operators literature, it is often convenient to have a symbolic representation for the pseudo-differential operator. First, the Fourier transform of a function $f$ is taken as $\hat{f}(\xi) = \int\nolimits_{\mathbb{R}^{n}}f(x)e^{-i2\pi\xi x}dx$. The pseudo-differential operator over a function $f$ is defined as
\begin{equation*}
    T_{a}f(x) = \int\nolimits_{\mathbb{R}^{n}}a(x, \xi)e^{i2\pi\xi x}\hat{f}(x)dx,
\end{equation*}
where the operator $T_{a}$ is parameterized by the symbol $a(x, \xi)$ which for the differential equation \eqref{apeqn:pseudoDiffCanonical} is given by
\begin{equation*}
    a(x, \xi) = \sum\nolimits_{\alpha\in\mathbb{A}}a_{\alpha}(x)(2\pi i\xi)^{\alpha}.
\end{equation*}
The Euler-Bernoulli equation discussed in the Section\,\ref{ssec:eulerBer} has $\mathbb{A} = \{0, 4\}$.

\section{Multiwavelet Filters}
\label{apsec:mwtFilters}
We discuss in details the multiwavelet filters as presented in the Section\,\ref{ssec:nsMWT}. First, we introduce some mathematical terminologies, specifically, useful for multiwavelets filters in Section\,\ref{apssec:measures}, and then preview few useful tools in Sections\,~\ref{apssec:gaussQuad},\,\ref{apssec:gso}.

\subsection{Measures, Basis, and Projections}
\label{apssec:measures}
\textbf{Measures:} The functions are expressed w.r.t. basis usually by using measures $\mu$ which could be non-uniform in-general. Intuitively, the measure provides weights to different locations over which the specified basis are defined. For a measure $\mu$, let us consider a Radon-Nikodym derivative as $w(x) \coloneqq \frac{d\mu}{d\lambda}(x)$, where, $d\lambda \coloneqq dx$ is the Lebesgue measure. In other words, the measure-dependent integrals $\int fd\mu(x)$, can now be defined as $\int f(x)w(x)dx$. 

\textbf{Basis:} A set of orthonormal basis w.r.t. measure $\mu$, are $\phi_{0}, \hdots, \phi_{k-1}$ such that $\langle \phi_{i}, \phi_{j}\rangle_{\mu} = \delta_{ij}$. With the weighting function $w(x)$, which is a Radon-Nikodym derivative w.r.t. Lebesgue measure, the orthonormality condition can be re-written as $\int\phi_{i}(x)\phi_{j}(x)w(x)dx = \delta_{ij}$.

The basis can also be appended with a multiplicative function called \textit{tilt} $\chi(x)$ such that for a set of basis $\phi_{i}$ which is orthonormal w.r.t. $\mu$ with weighting function $\frac{d\mu}{d\lambda}(x) = w(x)$, a new set of basis $\phi_{i}\chi$ are now orthonormal w.r.t. a measure having weighting function $w/\chi^2$. We will see that for OPs like Chebyshev in Section\,\ref{apssec:chebyshev}, a proper choice of tilt $\chi(x)$ simplifies the analysis.

\textbf{Projections:} For a given set of basis $\phi_{i}$ defined w.r.t. measure $\mu$ and corresponding weight function $w(x)$, the inner-products are defined such that they induce a measure-dependent Hilbert space structure $\mathcal{H}_{\mu}$. Next, for a given function $f$ such that $f\in\mathcal{H}_{\mu}$, the projections onto the basis polynomials are defined as $c_{i} = \int f(x)\phi_{i}(x)w(x)dx$. 

\subsection{Gaussian Quadrature}
\label{apssec:gaussQuad}

The Gaussian quadrature are the set of tools which are useful in approximating the definite integrals of the following form
\begin{equation}
    \int\nolimits_{a}^{b}f(x)w(x)dx \approx \sum\limits_{i=1}^{n}\omega_{i}f(x_{i}),
    \label{apeqn:gaussQuadOrig}
\end{equation}
where, $\omega_{i}$ are the scalar weight coefficients, and $x_{i}$ are the $n$ locations chosen appropriately. For a $n$-point quadrature, the eq. \eqref{apeqn:gaussQuadOrig} is exact for the functions $f$ that are polynomials of degree $\leq2n-1$. This is particularly useful to us, as we see in the Section\,\ref{apsec:mwtFiltersDerive}. 

From the result in \cite{stoer2002introduction}, it can be argued that, for a class of OPs $P_{i}$ defined w.r.t. weight function $w(x)$ over the interval $[a, b]$ such that $x_{1}, x_{2}, \hdots, x_{n}$ are the roots of $P_{n}$, if 
\begin{equation*}
    \sum\limits_{i=1}^{n}\omega_{i}P_{k}(x_{i}) = 
    \begin{cases}
    \vert\vert P_{0}\vert\vert_{\mu}^{2} & k = 0,\\
    0 & k>0,
    \end{cases}
\end{equation*}
then,
\begin{equation*}
    \sum\limits_{i = 1}^{n}\omega_{i}f(x_{i}) = \int\nolimits_{a}^{b}f(x)w(x)dx,
\end{equation*}
for any $f$ such that $f$ is a polynomial of degree $\leq 2n-1$. The weight coefficients can also be written in a closed-form expression \cite{abramowitz1965handbook} as follows
\begin{equation}
    \omega_{i} = \frac{a_{n}}{a_{n-1}}\frac{\int\nolimits_{a}^{b}P_{n-1}^{2}(x)w(x)dx}{P_{n}^{\prime}(x_{i})P_{n-1}(x_{i})},
    \label{apeqn:gaussQuadWeights}
\end{equation}
where, $a_{n}$ is the coefficient of $x^{n}$ in $P_{n}$. Thus, the integral in \eqref{apeqn:gaussQuadOrig} can be computed using family of OPs defined w.r.t. weight function $w(x)$. Depending on the class of OPs chosen, the Gaussian quadrature formula can be derived accordingly using eq. \eqref{apeqn:gaussQuadWeights}. For a common choice of OPs, the corresponding name for the Quadrature is 'Gaussian-Legendre', 'Gaussian-Chebyshev', 'Gaussian-Laguerre', etc.

\subsection{Gram-Schmidt Orthogonalization}
\label{apssec:gso}

The Gram-Schmidt Orthogonalization (GSO) is a common technique for deriving a (i) set of vectors in a subspace, orthogonal to an (ii) another given set of vectors. We briefly write the GSO procedure for obtaining a set of orthonormal polynomials w.r.t. measures which in-general is different for polynomials in set (i) and (ii). Specifically, we consider that for a given subspace of polynomials with degree $<k$ as $V_{0}$ and another subspace of polynomials with degree $<k$ $V_{1}$, such that $V_{0}\subset V_{1}$, we wish to obtain a set of orthonormal basis for the subspace of polynomials with degree $<k$ $W_{0}$, such that $V_{0}\perp W_{0}$ and $W_{0}\subset V_{1}$. It is apparent that, if dim$(W_{0})= n$,  dim$(V_{0}) = m$ and dim$(V_{1}) = p$, then $m+n\leq p$.

Let $(\psi_{0}, \hdots, \psi_{n-1})$ be a set of basis of the polynomial subspace $W_{0}$, $(\phi_{0}^{(0)},\hdots,\phi_{m-1}^{(0)})$ be a set of basis for $V_{0}$, and $(\phi_{1}^{(0)},\hdots,\phi_{p-1}^{(1)})$ be a set of basis for $V_{1}$. We take that basis $\psi_{i}$ and $\phi_{i}^{(0)}$ are defined w.r.t. same measure $\mu_{0}$, while $\phi_{i}^{(1)}$ are defined w.r.t. a different measure $\mu_{1}$. A set of $\psi_{i}$ can be obtained by iteratively applying the following procedure for $i=0,1,\hdots,n-1$
\begin{align}
\begin{aligned}
    \psi_{i} &\leftarrow \phi_{i}^{(1)} - \sum\nolimits_{j = 0}^{m-1}\langle \phi_{i}^{(1)}, \phi_{j}^{(0)}\rangle_{\mu_{0}}\phi_{j}^{(0)} - \sum\nolimits_{l = 0}^{i-1}\langle \phi_{i}^{(i)}, \psi_{l}\rangle_{\mu_{0}}\psi_{l},\\
    \psi_{i}  &\leftarrow \frac{\psi_{i}}{\vert\vert\psi_{i}\vert\vert_{\mu_{0}}}.
\end{aligned}
\label{apeqn:gso}
\end{align}
The procedure in \eqref{apeqn:gso} results in a set of orthonormal basis of $W_{0}$ such that $\langle \psi_{i}, \psi_{j}\rangle_{\mu_{0}} = \delta_{ij}$ as well as $\langle \psi_{i}, \phi_{j}^{(0)}\rangle_{\mu_{0}},\,\,\forall\, 0\leq i<n, 0\leq j<m$. We will see in Section\,\ref{apsec:mwtFiltersDerive} that the inner-product integrals in eq. \eqref{apeqn:gso} can be efficiently computed using the Gaussian Quadrature formulas (as discussed in Section\,\ref{apssec:gaussQuad}).

\section{Derivations for Multiwavelet Filters}
\label{apsec:mwtFiltersDerive}
Using the mathematical preliminaries and tools discussed in the Sections\,~\ref{apsec:techBack} and \,~\ref{apsec:mwtFilters}, we are now in shape to present a detailed derivations for the measure dependent multiwavelet filters. We start with deriving the general filters expressions in Section\,\ref{apssec:filtersProj}. Particular expressions for Legendre polynomials are presented in Section\,\ref{apssec:legendre}, and then for Chebyshev polynomials in Section\,\ref{apssec:chebyshev}.

\subsection{Filters as subspace projection coefficients}
\label{apssec:filtersProj}
The `multiwavelet filters' play the role of transforming the multiwavelet coefficients from one scale to another. Let us revisit the Section\,\ref{ssec:nsMWT}, where we defined a space of piecewise polynomial functions, for $k\in\mathbb{N}$ and $n\in\mathbb{Z}^{+}\cup\{0\}$ as, ${\bf{V}}_{n}^{k}$. The $\text{dim}({\bf V}_{n}^{k}) = 2^{n}k$, and for subsequent $n$, each subspace is contained in another, i.e, $V_{n-1}^{k}\subset V_{n}^{k}$. Now, if $\phi_{0}, \hdots,\phi_{k-1}$ are a set of basis polynomials for $V_{0}^{k}$ w.r.t. measure $\mu_{0}$, then we know that a set of basis for $V_{1}^{k}$ can be obtained by scale and shift of $\phi_{i}$ as $\phi_{jl}^{1} = 2^{1/2}\phi_{j}(2x-l)\,\,l = 0, 1$, and the measure accordingly as $\mu_{1}$. For a given function $f$, its multiwavelet coefficients for projections over $V_{0}^{k}$ are taken as $s_{0i}^{0} = \langle f, \phi_{i}\rangle){\mu_{0}}$ and for $V_{1}^{k}$ is taken as $s_{li}^{1} = \langle f, \phi_{il}^{1}\rangle_{\mu_{1}}$, and, we are looking for filter coefficients $(H)$ such that a transformation between projections at these two consecutive scale exists, or
\begin{equation}
    s_{0i}^{0} = \sum\nolimits_{l=0,1}\sum\nolimits_{j=0}^{k-1}H_{ij}^{(l)}s_{lj}^{1}.
    \label{apeqn:filter0scale}
\end{equation}
Let us begin by considering a simple scenario. Since, $V_{0}^{k}\subset V_{1}^{k}$, the basis are related as
\begin{equation}
    \phi_{i} = \sum\nolimits_{j = 0}^{k-1}\alpha_{ij}^{(0)}\sqrt{2}\phi_{j}(2x) + \sum\nolimits_{j = 0}^{k-1}\alpha_{ij}^{(1)}\sqrt{2}\phi_{j}(2x-1).
    \label{apeqn:subspaceCoeff}
\end{equation}
It is straightforward to see that if $\phi_{i}$ and $\phi_{il}^{1}$ are defined w.r.t. same measure, or $\mu_{0} = \mu_{1}$ almost everywhere (a.e.), then the filters transforming the multiwavelet coefficients from higher to lower scale, are exactly equal to the subspace mapping coefficients $\alpha_{ij}^{(0)}, \alpha_{ij}^{(1)}$ ( by taking inner-product with $f$ on both sides in \eqref{apeqn:subspaceCoeff}). However, this is not the case in-general, i.e., the measures w.r.t. which the basis are defined at each scale are not necessarily same. To remedy this issue, and to generalize the multiwavelet filters, we now present a general measure-variant version of the multiwavelet filters.

We note that, solving for filters $H$ that satisfy  eq. \eqref{apeqn:filter0scale} indeed solves the general case of $n+1\rightarrow n$ scale, which can be obtained by a simple change of variables as $s_{l,i}^{n} = \sum\nolimits_{j=0}^{k-1}H_{ij}^{(0)}s_{2l,j}^{n+1} + \sum\nolimits_{j=0}^{k-1}H_{ij}^{(1)}s_{2l+1,j}^{n+1}$. Now, for solving \eqref{apeqn:filter0scale}, we consider the following equation
\begin{equation}
    \phi_{i}(x)\frac{d\mu_{0}}{d\lambda}(x) = \sum\limits_{j=0}^{k-1}H_{ij}^{(0)}\sqrt{2}\phi_{j}(2x)\frac{d\mu_{1}}{d\lambda}(x) + \sum\limits_{j=0}^{k-1}H_{ij}^{(1)}\sqrt{2}\phi_{j}(2x-1)\frac{d\mu_{1}}{d\lambda}(x),
    \label{apeqn:measureTransform}
\end{equation}
where $\frac{d\mu}{d\lambda}$ is the Radon-Nikodym derivative as discussed in Section\,\ref{apssec:measures}, and we have also defined $d\lambda \coloneqq dx$. We observe that eq. \eqref{apeqn:filter0scale} can be obtained from \eqref{apeqn:measureTransform} by simply integrating with $f$ on both sides. 

Next, we observe an important fact about multiwavelets (or wavelets in-general) that the advantages offered by multiwavelets rely on their ability to project a function \textit{locally}. One way to achieve this is by computing basis functions which are dilation/translations of a fixed mother wavelet, for example, Figure\,\ref{apfig:wavelDemo}. However, the idea can be generalized by projecting a given function onto any set of basis as long as they capture the \textit{locality}. A possible approach to generalize is by using a tilt variant of the basis at higher scales, i.e., using $\sqrt{2}\Tilde{\phi}_{i}(2x)=\sqrt{2}\phi_{i}(2x)\chi_{0}(x)$, and $\sqrt{2}\Tilde{\phi}_{i}(2x-1)=\sqrt{2}\phi_{i}(2x-1)\chi_{1}(x)$ such that $\sqrt{2}\Tilde{\phi}_{i}(2x)$ are now orthonormal w.r.t. weighting function $w(2x)/\chi_{0}^{2}(x)$, and similarly $\sqrt{2}\Tilde{\phi}_{i}(2x-1)$ w.r.t. $w(2x-1)/\chi_{1}^{2}(x)$. By choosing $\chi_{0}(x) = w(2x)/w(x)$, and $\chi_{1}(x) = w(2x-1)/w(x)$, and taking the new tilted measure $\Tilde{\mu}_{1}$ such that 
\begin{align*}
    \Tilde{\mu}_{1}([0,1]) = \int\nolimits_{0}^{1/2}\frac{w(2x)}{\chi_{0}^{2}(x)}d\lambda(x) + \int\nolimits_{1/2}^{1}\frac{w(2x-1)}{\chi_{1}^{2}(x)}d\lambda(x),
\end{align*}
or,
\begin{align*}
    \frac{d\Tilde{\mu}_{1}}{d\lambda}(x) = 
    \begin{cases}
        \frac{w(2x)}{\chi_{0}^{2}(x)} & 0\leq x\leq 1/2,\\
        \frac{w(2x-1)}{\chi_{1}^{2}(x)} & 1/2< x\leq 1.\\
    \end{cases}
\end{align*}
We re-write the eq. \eqref{apeqn:measureTransform}, by substituting $\phi_{i}(2x)\leftarrow\Tilde{\phi}_{i}(2x), \phi_{i}(2x-1)\leftarrow\Tilde{\phi}_{i}(2x-1)$ and $\mu_{1}\leftarrow\Tilde{\mu}_{1}$, in its most useful form for the current work as follows
\begin{equation*}
    \phi_{i}(x)w(x) = \sum\limits_{j=0}^{k-1}H_{ij}^{(0)}\sqrt{2}\phi_{j}(2x)w(x) + \sum\limits_{j=0}^{k-1}H_{ij}^{(1)}\sqrt{2}\phi_{j}(2x-1)w(x),
\end{equation*}
or,
\begin{equation}
    \phi_{i}(x) = \sum\limits_{j=0}^{k-1}H_{ij}^{(0)}\sqrt{2}\phi_{j}(2x) + \sum\limits_{j=0}^{k-1}H_{ij}^{(1)}\sqrt{2}\phi_{j}(2x-1), \qquad (a.e.).
    \label{apeqn:measureTransformUse}
\end{equation}
Thus, \textit{filter coefficients can be looked upon as subspace projection coefficients}, with a proper choice of \textit{tilted} basis. Note that eq.\eqref{apeqn:measureTransformWavelUse} is now equivalent to \eqref{apeqn:subspaceCoeff} but is an outcome of a different back-end machinery. Since, $\sqrt{2}\phi_{i}(2x), \sqrt{2}\phi_{i}(2x-1)$ are orthonormal basis for $V_{1}^{k}$, we have 
\begin{equation*}
    2\int\nolimits_{0}^{1/2}\phi_{i}(2x)\phi_{j}(2x)w(2x)dx = \delta_{ij},\quad 2\int\nolimits_{1/2}^{1}\phi_{i}(2x-1)\phi_{j}(2x-1)w(2x-1)dx = \delta_{ij},
\end{equation*}
and hence we obtain the filter coefficients as follows
\begin{align}
    H_{ij}^{(0)} &= \sqrt{2}\int\nolimits_{0}^{1/2}\phi_{i}(x)\phi_{j}(2x)w(2x)dx,\label{apeqn:filterH0}\\
    H_{ij}^{(1)} &= \sqrt{2}\int\nolimits_{1/2}^{1}\phi_{i}(x)\phi_{j}(2x-1)w(2x-1)dx.\label{apeqn:filterH1}
\end{align}
For a given set of basis of $V_{0}^{k}$ as $\phi_{0},\hdots,\phi_{k-1}$ defined w.r.t. measure/weight function $w(x)$, the filter coefficients $H$ can be derived by solving eq. \eqref{apeqn:measureTransformUse}. In a similar way, if $\psi_{0},\hdots,\psi_{k-1}$ is the basis for the multiwavelet subspace $W_{0}^{k}$ w.r.t. measure $\mu_{0}$ such that $V_{0}^{k} \bigoplus W_{0}^{k} = V_{1}^{k}$, and the projection of function $f$ over $W_{0}^{k}$ is denoted by $d_{0,i}^{0} = \langle f, \psi_{i}\rangle_{\mu_{0}}$, then the filter coefficients for obtaining the multiwavelet coefficients is written as
\begin{equation}
    d_{0i}^{0} = \sum\nolimits_{l=0,1}\sum\nolimits_{j=0}^{k-1}G_{ij}^{(l)}s_{lj}^{1}.
    \label{apeqn:filter0wavel}
\end{equation}
Again using a change of variables, we get $d_{l,i}^{n} = \sum\nolimits_{j=0}^{k-1}G_{ij}^{(0)}s_{2l,j}^{n+1} + \sum\nolimits_{j=0}^{k-1}G_{ij}^{(1)}s_{2l+1,j}^{n+1}$. To solve for $G$ in \eqref{apeqn:filter0wavel}, similar to eq. \eqref{apeqn:measureTransformUse}, the measure-variant multiwavelet basis transformation (with appropriate tilt) is written as
\begin{equation}
    \psi_{i}(x) = \sum\limits_{j=0}^{k-1}G_{ij}^{(0)}\sqrt{2}\phi_{j}(2x) + \sum\limits_{j=0}^{k-1}G_{ij}^{(1)}\sqrt{2}\phi_{j}(2x-1), \qquad (a.e.).
    \label{apeqn:measureTransformWavelUse}
\end{equation}
Similar to eq. \eqref{apeqn:filterH0}-\eqref{apeqn:filterH1}, the filter coefficients $G$ can be obtained from \eqref{apeqn:measureTransformWavelUse} as follows
\begin{align}
    G_{ij}^{(0)} &= \sqrt{2}\int\nolimits_{0}^{1/2}\psi_{i}(x)\phi_{j}(2x)w(2x)dx,\label{apeqn:filterG0}\\
    G_{ij}^{(1)} &= \sqrt{2}\int\nolimits_{1/2}^{1}\psi_{i}(x)\phi_{j}(2x-1)w(2x-1)dx.\label{apeqn:filterG1}
\end{align}
Since $\langle \phi_{i}, \phi_{j}\rangle_{\mu_{0}} = \delta_{ij}$, $\langle \psi_{i}, \psi_{j}\rangle_{\mu_{0}} = \delta_{ij}$ and $\langle \phi_{i}, \psi_{j}\rangle_{\mu_{0}} = 0$, therefore, using \eqref{apeqn:measureTransformUse}, \eqref{apeqn:measureTransformWavelUse}, we can write that
\begin{align}
    \int\nolimits_{0}^{1}\phi_{i}(x)\phi_{j}(x)w(x)dx &= 2\sum\limits_{l=0}^{k-1}\sum\limits_{l^{\prime}=0}^{k-1}H_{il}^{(0)}H_{jl^{\prime}}^{(0)}\int\nolimits_{0}^{1/2}\phi_{l}(2x)\phi_{l^{\prime}}(2x)w(x)dx\nonumber\\
     &\quad+ 2\sum\limits_{l=0}^{k-1}\sum\limits_{l^{\prime}=0}^{k-1}H_{il}^{(1)}H_{jl^{\prime}}^{(1)}\int\nolimits_{1/2}^{1}\phi_{l}(2x-1)\phi_{l^{\prime}}(2x-1)w(x)dx,
\end{align}
\begin{align}
    \int\nolimits_{0}^{1}\psi_{i}(x)\psi_{j}(x)w(x)dx &= 2\sum\limits_{l=0}^{k-1}\sum\limits_{l^{\prime}=0}^{k-1}G_{il}^{(0)}G_{jl^{\prime}}^{(0)}\int\nolimits_{0}^{1/2}\phi_{l}(2x)\phi_{l^{\prime}}(2x)w(x)dx\nonumber\\
     &\hspace*{20pt}+ 2\sum\limits_{l=0}^{k-1}\sum\limits_{l^{\prime}=0}^{k-1}G_{il}^{(1)}G_{jl^{\prime}}^{(1)}\int\nolimits_{1/2}^{1}\phi_{l}(2x-1)\phi_{l^{\prime}}(2x-1)w(x)dx,
\end{align}
\begin{align}
    0 &= 2\sum\limits_{l=0}^{k-1}\sum\limits_{l^{\prime}=0}^{k-1}H_{il}^{(0)}G_{jl^{\prime}}^{(0)}\int\nolimits_{0}^{1/2}\phi_{l}(2x)\phi_{l^{\prime}}(2x)w(x)dx\nonumber\\
     &\quad+ 2\sum\limits_{l=0}^{k-1}\sum\limits_{l^{\prime}=0}^{k-1}H_{il}^{(1)}G_{jl^{\prime}}^{(1)}\int\nolimits_{1/2}^{1}\phi_{l}(2x-1)\phi_{l^{\prime}}(2x-1)w(x)dx.
\end{align}
Let us define filter matrices as $H^{(l)} = [H_{ij}^{(l)}]\in\mathbb{R}^{k\times k}$ and $G^{(l)} = [G_{ij}^{(l)}]\in\mathbb{R}^{k\times k}$ for $l=0,1$. Also, we define correction matrices as $\Sigma^{(0)} = [\Sigma_{ij}^{(0)}], \Sigma^{(1)} = [\Sigma_{ij}^{(1)}]$ such that 
\begin{align}
    \begin{aligned}
        \Sigma_{ij}^{(0)} &= 2\int\nolimits_{0}^{1/2}\phi_{i}(2x)\phi_{j}(2x)w(x)dx,\\ 
        \Sigma_{ij}^{(1)} &= 2\int\nolimits_{1/2}^{1}\phi_{i}(2x-1)\phi_{j}(2x-1)w(x)dx.
    \end{aligned}
    \label{apeqn:correctionMatrices}
\end{align}
Now, we can write that
\begin{align}
    \begin{aligned}
        H^{(0)}\Sigma^{(0)}H^{(0)\,T} + H^{(1)}\Sigma^{(1)}H^{(1)\,T} &= I,\\
        G^{(0)}\Sigma^{(0)}G^{(0)\,T} + G^{(1)}\Sigma^{(1)}G^{(1)\,T} &= I,\\
        H^{(0)}\Sigma^{(0)}G^{(0)\,T} + H^{(1)}\Sigma^{(1)}G^{(1)\,T} &= 0.
    \end{aligned}
\end{align}
Rearranging eq. we can finally express the relationships between filter matrices and correction matrices as follows
\begin{equation}
    \begin{bmatrix}
    H^{(0)} & H^{(1)}\\
    G^{(0)} & G^{(1)}
    \end{bmatrix}
    \begin{bmatrix}
    \Sigma^{(0)} & 0\\
    0 & \Sigma^{(1)}
    \end{bmatrix}
    \begin{bmatrix}
    H^{(0)} & H^{(1)}\\
    G^{(0)} & G^{(1)}
    \end{bmatrix}^{T}=I.
    \label{apeqn:filterConstr}
\end{equation}
The discussion till now is related to `decomposition' or transformation of multiwavelet transform coefficients from higher to lower scale. However, the other direction, i.e., `reconstruction' or transformation from lower to higher scale can also be obtained from \eqref{apeqn:filterConstr}. First, note that the general form of eq. \eqref{apeqn:filter0scale}, \eqref{apeqn:filter0wavel} can be written in the matrix format as 
\begin{align}
    \begin{aligned}
        {\bf s}_{l}^{n} &= H^{(0)}{\bf s}_{2l}^{n+1} + H^{(1)}{\bf s}_{2l+1}^{n+1},\\
        {\bf d}_{l}^{n} &= G^{(0)}{\bf s}_{2l}^{n+1} + G^{(1)}{\bf s}_{2l+1}^{n+1}.
    \end{aligned}
    \label{apeqn:coeffMatrix}
\end{align}
Next, we observe that $\Sigma^{(0)}, \Sigma^{(1)}\succ 0$, which follows from their definition. Therefore, eq. \eqref{apeqn:filterConstr} can be inverted to get the following form
\begin{equation}
    \begin{bmatrix}
    H^{(0)} & H^{(1)}\\
    G^{(0)} & G^{(1)}
    \end{bmatrix}
    \begin{bmatrix}
    H^{(0)} & H^{(1)}\\
    G^{(0)} & G^{(1)}
    \end{bmatrix}^{T}=\begin{bmatrix}
    \Sigma^{(0)\,-1} & 0\\
    0 & \Sigma^{(1)\,-1}
    \end{bmatrix}.
    \label{apeqn:filterConstr1}
\end{equation}
Finally, by using \eqref{apeqn:filterConstr1}, we can essentially invert the eq. \eqref{apeqn:coeffMatrix} to get
\begin{align}
    \begin{aligned}
        {\bf s}_{2l}^{n+1} = \Sigma^{(0)}(H^{(0)\,T}{\bf s}_{l}^{n} + G^{(0)\,T}{\bf d}_{l}^{n}),\\
        {\bf s}_{2l+1}^{n+1} = \Sigma^{(1)}(H^{(1)\,T}{\bf s}_{l}^{n} + G^{(1)\,T}{\bf d}_{l}^{n}).
    \end{aligned}
    \label{apeqn:evenOdd}
\end{align}
In the following Section\,\ref{apssec:legendre},\ref{apssec:chebyshev} we see the the filters $H, G$ in \eqref{apeqn:coeffMatrix}, \eqref{apeqn:evenOdd} for different polynomial basis.

\subsection{Multiwavelets using Legendre Polynomials}
\label{apssec:legendre}
The basis for $V_{0}^{k}$ are chosen as normalized shifted Legendre polynomials of degree upto $k$ w.r.t. weight function $w_{L}(2x-1) = {\bf 1}_{[0,1]}(x)$ from Section\,\ref{apsssec:legendre}. For example, the first three bases are 
\begin{align}
    \begin{aligned}
        \phi_{0}(x) &= 1,\\
        \phi_{1}(x) &= \sqrt{3}(2x-1),\\
        \phi_{2}(x) &= \sqrt{5}(6x^2-6x+1),\quad 0\leq x \leq 1.
    \end{aligned}
\end{align}

For deriving a set of basis $\psi_{i}$ of $W_{0}^{k}$ using GSO, we need to evaluate the integrals which could be done efficiently using Gaussian quadrature.

\textbf{Gaussian-Legendre Quadrature:} The integrals involved in GSO procedure, and the computations of $H, G$ can be done efficiently using the Gaussian quadrature as discussed in Section\,\ref{apssec:gaussQuad}. Since the basis functions $\phi_{i}, \psi_{i}$ are polynomials, therefore, the quadrature summation would be \textit{exact}. For a given $k$ basis of the subspace $V_{0}^{k}$, the deg $(\phi_{i}\phi_{j})<2k-1$, as well as deg $(\phi_{i}\psi_{j})<2k-1$, therefore a $k$-point quadrature would be sufficient for expressing the integrals. Next, we take the interval $[a, b] = [0, 1]$, and the OPs for approximation in Gaussian quadrature as shifted Legendre polynomials $P_{k}(2x-1)$. The weight coefficients $\omega_{i}$ can be written as 
\begin{align}
    \omega_{i} &= \frac{a_{k}}{a_{k-1}}\frac{\int\nolimits_{0}^{1}P_{k-1}^{2}(2x-1)w(2x-1)dx}{P_{k}^{\prime}(2x_{i}-1)P_{k-1}(2x_{i}-1)}\nonumber\\
    &= \frac{2k-1}{k}.\frac{1}{2k-1}\frac{1}{P_{k}^{\prime}(2x_{i}-1)P_{k-1}(2x_{i}-1)} = \frac{1}{kP_{k}^{\prime}(2x_{i}-1)P_{k-1}(2x_{i}-1)},
\end{align}
where $x_{i}$ are the $k$ roots of $P_{k}(2x-1)$ and $a_{k}$ can be expressed in terms of $a_{k-1}$ using the recurrence relationship of Legendre polynomials from Section\,\ref{apsssec:legendre}.

A set of basis for $V_{1}^{k}$ is $\sqrt{2}\phi_{i}(2x)$ and $\sqrt{2}\phi_{i}(2x-1)$ with weight functions $w_{L}(4x-1) = {\bf 1}_{[0,1/2]}(x)$ and $w_{L}(4x-3) = {\bf 1}_{(1/2,1]}(x)$, respectively. We now use GSO procedure as outlined in Section\,\ref{apssec:gso} to obtain set of basis $\psi_{0},\hdots,\psi_{k-1}$ for $W_{0}^{k}$. We use Gaussian-Legendre quadrature formulas for computing the inner-products. As an example, the inner-products are computed as follows
\begin{align*}
    \langle \sqrt{2}\phi_{i}, \phi_{j}\rangle_{\mu_{0}} &= \int\nolimits_{0}^{1}\sqrt{2}\phi_{i}(2x)\phi_{j}(x)w_{L}(2x-1)dx\\
    &= {\sqrt{2}}\sum\limits_{i = 1}^{k}\omega_{i}\phi_i(2x_i)\phi_{j}({x_i}),
\end{align*}
where $\phi_{i}(2x_i) = 0$ for $x_{i}>0.5$.

With shifted Legendre polynomials as basis for $V_{0}^{3}$, the multiwavelet bases for $W_{0}^{3}$ are
\begin{align}
    \begin{aligned}
        \psi_{0}(x) &= 
        \begin{cases}
            6x - 1 & 0\leq x\leq 1/2,\\
            6x - 5 & 1/2< x\leq 1,
        \end{cases}\\
        \psi_{1}(x) &= 
        \begin{cases}
            \sqrt{3}(30x^2 - 14x + 1) & 0\leq x\leq 1/2,\\
            \sqrt{3}(30x^2 - 46x + 17) & 1/2< x\leq 1,
        \end{cases}\\
        \psi_{2}(x) &= 
        \begin{cases}
            \sqrt{5}(24x^2 - 12x + 1) & 0\leq x\leq 1/2,\\
            \sqrt{5}(-24x^2 + 36x - 13) & 1/2< x\leq 1.
        \end{cases}
    \end{aligned}
\end{align}

Next, we compute the filter matrices, but first note that since the weighting function for Legendre polynomials basis are $w_{L}(x) = {\bf 1}_{[0,1]}(x)$, therefore, $\Sigma^{(0)}, \Sigma^{(1)}$ in eq. \eqref{apeqn:correctionMatrices} are just identity matrices because of orthonormality of the basis $\sqrt{2}\phi_{i}(2x)$ and $\sqrt{2}\phi_{i}(2x-1)$ w.r.t. ${\bf 1}_{[0,1/2]}(x)$ and ${\bf 1}_{[1/2,1]}(x)$, respectively. The filter coefficients can be computed using Gaussian-Legendre quadrature as follows
\begin{align*}
    H_{ij}^{(0)} &= \sqrt{2}\int\nolimits_{0}^{1/2}\phi_{i}(x)\phi_{j}(2x)w_{L}(2x-1)dx\\
    &= \frac{1}{\sqrt{2}}\int\nolimits_{0}^{1}\phi_{i}(x/2)\phi_j(x)dx\\
    &=\frac{1}{\sqrt{2}}\sum\limits_{i = 1}^{k}\omega_{i}\phi_i\left(\frac{x_i}{2}\right)\phi_{j}(x_i),
\end{align*}
and similarly other coefficients can be obtained in eq. \eqref{apeqn:filterH0}-\eqref{apeqn:filterH1}, \eqref{apeqn:filterG0}-\eqref{apeqn:filterG1}. As an example, for $k=3$, following the outlined procedure, the filter coefficients are derived as follows
\begin{align*}
    H^{(0)} &= 
    \begin{bmatrix}
        \frac{1}{\sqrt{2}} & 0 & 0 \\
        -\frac{\sqrt{3}}{2\sqrt{2}} & \frac{1}{2\sqrt{2}} & 0 \\
        0 & -\frac{\sqrt{15}}{4\sqrt{2}} & \frac{1}{4\sqrt{2}}
    \end{bmatrix},
    \qquad
    &H^{(1)} &= 
    \begin{bmatrix}
        \frac{1}{\sqrt{2}} & 0 & 0 \\
        \frac{\sqrt{3}}{2\sqrt{2}} & \frac{1}{2\sqrt{2}} & 0 \\
        0 & \frac{\sqrt{15}}{4\sqrt{2}} & \frac{1}{4\sqrt{2}}
    \end{bmatrix},
    \\
    G^{(0)} &= 
    \begin{bmatrix}
        \frac{1}{2\sqrt{2}} &  \frac{\sqrt{3}}{2\sqrt{2}} & 0 \\
        0 & \frac{1}{4\sqrt{2}} & \frac{\sqrt{15}}{4\sqrt{2}} \\
        0 & 0 & \frac{1}{\sqrt{2}}
    \end{bmatrix},
    \qquad
    &G^{(1)} &= 
    \begin{bmatrix}
        -\frac{1}{2\sqrt{2}} &  \frac{\sqrt{3}}{2\sqrt{2}} & 0 \\
        0 & -\frac{1}{4\sqrt{2}} & \frac{\sqrt{15}}{4\sqrt{2}} \\
        0 & 0 & -\frac{1}{\sqrt{2}}
    \end{bmatrix}.
\end{align*}

\subsection{Multiwavelets using Chebyshev Polynomials}
\label{apssec:chebyshev}
We choose the basis for $V_{0}^{k}$ as shifted Chebyshev polynomials of the first-order from degree $0$ to $k-1$. The weighting function for shifted Chebyshev polynomials is $w_{Ch}(2x-1) = 1\sqrt{1-(2x-1)^2}$ from Section\,\ref{apsssec:chebyshev}. The first three bases using Chebyshev polynomials are as follows
\begin{align*}
    \phi_{0}(x) &= \sqrt{2/\pi},\\
    \phi_{1}(x) &= \frac{2}{\sqrt{\pi}}(2x-1),\\
    \phi_{2}(x) &= \frac{2}{\sqrt{\pi}}(8x^2 - 8x  +1), 0\leq x\leq 1.
\end{align*}

The Gaussian quadrature for the Chebyshev polynomials is used to evaluate the integrals that appears in the GSO procedure as well as in the computations of filters $H, G$.

\textbf{Gaussian-Chebyshev Quadrature:} The basis functions $\phi_{i}, \psi_{i}$ resulting from the use of shifted Chebyshev polynomials are also polynomials with degree of their products such that deg$(\phi_{i}\phi_{j})<2k-1$ and deg$(\phi_{i}\psi_{i})<2k-1$, therefore a $k$-point quadrature would be sufficient for evaluating the integrals that have products of bases. Upon taking the interval $[a, b]$ as $[0, 1]$, and using the canonical OPs as shifted Chebyshev polynomials, the weight coefficients are written as 
\begin{align}
    \omega_{i} &= \frac{a_{k}}{a_{k-1}}\frac{\int\nolimits_{0}^{1}T_{k-1}^{2}(2x-1)w_{Ch}(2x-1)dx}{T_{k}^{\prime}(2x_{i}-1)T_{k-1}(2x_{i}-1)}\nonumber\\
    &\stackrel{(a)}{=}2\frac{\pi}{4}\frac{1}{T_{k}^{\prime}(2x_{i}-1)T_{k-1}(2x_{i}-1)}\nonumber\\ 
    &\stackrel{(b)}{=}\frac{\pi}{2k},
\end{align}
where $x_{i}$ are the $k$ roots of $T_{k}(2x-1)$, $(a)$ is using the fact that $a_{n}/a_{n-1} = 2$ by using the recurrence relationship of Chebyshev polynomials from Section\,\ref{apsssec:chebyshev}, and assumes $k>1$ for the squared integral. For $(b)$, we first note that $T_{k}(\cos\theta) = \cos(k\theta)$, hence, $T_{k}^{\prime}(\cos\theta) = n\sin(n\theta)/\sin(\theta)$. Since $x_{i}$ are the roots of $T_{k}(2x-1)$, therefore, $2x_{i}-1 = \cos(\frac{\pi}{n}(i-1/2))$. Substituting the $x_{i}$, we get $T_{k}^{\prime}(2x_{i}-1)T_{k-1}(2x_i-1) = k$.

A set of basis for $V_{1}^{k}$ is $\sqrt{2}\phi_{i}(2x)$ and $\sqrt{2}\phi_{i}(2x-1)$ with weight functions $w_{Ch}(4x-1) = 1/\sqrt{1-(4x-1)^2}$ and $w_{Cb}(4x-3) = 1/\sqrt{1 - (4x-3)^2}$, respectively. We now use GSO procedure as outlined in Section\,\ref{apssec:gso} to obtain set of basis $\psi_{0},\hdots,\psi_{k-1}$ for $W_{0}^{k}$. We use Gaussian-Chebyshev quadrature formulas for computing the inner-products. As an example, the inner-products are computed as follows
\begin{align*}
    \langle \sqrt{2}\phi_{i}, \phi_{j}\rangle_{\mu_{0}} &= \int\nolimits_{0}^{1}\sqrt{2}\phi_{i}(2x)\phi_{j}(x)w_{Ch}(2x-1)dx\\
    &= {\sqrt{2}}\frac{\pi}{2k}\sum\limits_{i = 1}^{k}\phi_i(2x_i)\phi_{j}({x_i}),
\end{align*}
where $\phi_{i}(2x_i) = 0$ for $x_{i}>0.5$.

With shifted Chebyshev polynomials as basis for $V_{0}^{3}$, the multiwavelet bases for $W_{0}^{3}$ are derived as
\begin{align}
    \begin{aligned}
        \psi_{0}(x) &= 
        \begin{cases}
            4.9749x - 0.5560 & 0\leq x\leq 1/2,\\
            4.9749x - 4.4189 & 1/2< x\leq 1,
        \end{cases}\\
        \psi_{1}(x) &= 
        \begin{cases}
            58.3516x^2 -22.6187x + 0.9326 & 0\leq x\leq 1/2,\\
            58.3516x^2 -94.0846x + 36.6655 & 1/2< x\leq 1,
        \end{cases}\\
        \psi_{2}(x) &= 
        \begin{cases}
            59.0457x^2 -23.7328x + 1.0941 & 0\leq x\leq 1/2,\\
            -59.0457x^2 + 94.3586x -36.4070 & 1/2< x\leq 1.
        \end{cases}
    \end{aligned}
\end{align}
Next, we compute the filter and the correction matrices. The filter coefficients can be computed using Gaussian-Chebyshev quadrature as follows
\begin{align*}
    H_{ij}^{(0)} &= \sqrt{2}\int\nolimits_{0}^{1/2}\phi_{i}(x)\phi_{j}(2x)w_{Ch}(4x-1)dx\\
    &= \frac{1}{\sqrt{2}}\int\nolimits_{0}^{1}\phi_{i}(x/2)\phi_j(x)w_{Ch}(2x-1)dx\\
    &=\frac{\pi}{2\sqrt{2}k}\sum\limits_{i = 1}^{k}\phi_i\left(\frac{x_i}{2}\right)\phi_{j}(x_i),
\end{align*}
and similarly, other coefficients can be obtained in eq. \eqref{apeqn:filterH0}-\eqref{apeqn:filterH1}, \eqref{apeqn:filterG0}-\eqref{apeqn:filterG1}. Using the outlined procedure for Chebyshev based OP basis, for $k=3$, the filter and the corrections matrices are derived as
\begin{align*}
    H^{(0)} &= 
    \begin{bmatrix}
        \frac{1}{\sqrt{2}} & 0 & 0 \\
        -\frac{1}{2} & \frac{1}{2\sqrt{2}} & 0 \\
        -\frac{1}{4} & -\frac{1}{\sqrt{2}} & \frac{1}{4\sqrt{2}}
    \end{bmatrix},
    \qquad
    &H^{(1)} &= 
    \begin{bmatrix}
        \frac{1}{\sqrt{2}} & 0 & 0 \\
        \frac{1}{2} & \frac{1}{2\sqrt{2}} & 0 \\
        -\frac{1}{4} & \frac{1}{\sqrt{2}} & \frac{1}{4\sqrt{2}}
    \end{bmatrix},
    \\
    G^{(0)} &= 
    \begin{bmatrix}
        0.6094 & 0.7794 & 0\\
        0.6632 &  1.0272 & 1.1427\\
        0.6172& 0.9070& 1.1562
    \end{bmatrix},
    \qquad
    &G^{(1)} &= 
    \begin{bmatrix}
        -0.6094 & 0.7794 & 0\\
        0.6632 &  -1.0272 & 1.1427\\
        -0.6172& 0.9070& -1.1562
    \end{bmatrix},
    \\
    \Sigma^{(0)} &= 
    \begin{bmatrix}
        1        & -0.4071& -0.2144\\
        -0.4071&  0.8483& -0.4482\\
       -0.2144& -0.4482&  0.8400
    \end{bmatrix},
    \qquad
    &\Sigma^{(1)} &= 
    \begin{bmatrix}
        1        & 0.4071& -0.2144\\
        0.4071&  0.8483& 0.4482\\
       -0.2144& 0.4482&  0.8400
    \end{bmatrix}.
\end{align*}

\subsection{Numerical Considerations}
\label{apssec:numericalIssues}
The numerical computations of the filter matrices are done using Gaussian quadrature as discussed in Sections\,~\ref{apssec:legendre} and\,~\ref{apssec:chebyshev} for Legendre and Chebyshev polynomials, respectively. For odd $k$, a root of the canonical polynomial (either Legendre, Chebyshev) would be exactly $0.5$. Since the multiwavelets bases $\psi_{i}$ for $W_{0}^{k}$ are discontinuous at $0.5$, the quadrature sum can lead to an unexpected result due to the finite-precision of the roots $x_{i}$. One solution for this is to add a small number $\epsilon, \epsilon>0$ to $x_{i}$ to avoid the singularity. Another solution, which we have used, is to perform a $\Tilde{k}$-quadrature, where $\Tilde{k} = 2k$. Note that, any high value of quadrature sum would work as long as it is greater than $k$, and we choose an even value to avoid the root at the singularity ($x=0.5$).

To check the validity of the numerically computed filter coefficients from the Gaussian quadrature, we can use eq. \eqref{apeqn:filterConstr}. In a $k$-point quadrature, the summation involves up to $k$ degree polynomials, and we found that for large values of $k$, for example, $k>20$, the filter matrices tend to diverge from the mathematical constraint of \eqref{apeqn:filterConstr}. Note that this is not due to the involved mathematics but the precision offered by floating-point values. For the current work, we found values of $k$ in the range of $[1, 6]$ to be most useful. However, a future research should look into other possible alternatives to work around the numerical errors due to the floating-point precision.

\section{Additional Results}
\label{apsec:additRes}
We present numerical evaluation of the proposed multiwavelets-based models on an additional dataset of Navier-Stokes in Section\,\ref{apssec:nsEqn}. Next, in Section\,\ref{apssec:highRes}, we present numerical results for prediction at finer resolutions with the use of lower-resolution trained models. Section\,\ref{apssec:pseudoDiff} presents additional results on the evaluation of multiwavelets on pseudo-differential equations.

\subsection{Navier-Stokes Equation}
\label{apssec:nsEqn}
Navier-Stokes Equations \cite{acheson1991elementary,batchelor2000introduction} describe the motion of viscous fluid substances, which can be used to model the ocean currents, the weather, and air flow. We experiment on the 2-d Navier-Stokes equation for a viscous, incompressible fluid in vorticity form on the unit torus, where it takes the following form:
\begin{align}
    \begin{aligned}
        \frac{\partial w(x,t)}{\partial t}+u(x,t)\cdot \nabla w(x,t)&-\nu \Delta w(x,t)=f(x), & x\in (0,1)^2, t\in (0,T]\\
        \nabla \cdot u(x,t)&=0, &x\in (0,1)^2, t\in [0,T]\\
        w_0(x)&=w(x,t=0), &x\in (0,1)^2
    \end{aligned}
\end{align}
We set the experiments to learn the operator mapping the vorticity $w$ up to time 10 to $w$ at a later time $T>10$. More specifically, task for the neural operator is to map the first $T$ time units to last $T-10$ time units of vorticity $w$. To compare with the state-of-the-art model FNO \cite{li2020fourier} and other configurations under the same conditions, we use the same Navier-Stokes' data and the results that have been published in \cite{li2020fourier}. The initial condition is sampled as Gaussian random fields where  $w_{0}\sim \mathcal N(0,7^{\frac{3}{2}}(-\Delta+7^2I)^{-2.5})$ with periodic boundary conditions. The forcing function $f(x)=0.1(\sin(2\pi (x_1+x_2))+\cos(2\pi (x_1+x_2)))$. The experiments are conducted with \textcircled{1} the viscosities $\nu = 1e-3$, the final time $T=50$, the number of training pairs $N=1000$; \textcircled{2} $\nu = 1e-4, T=30, N=1000$; \textcircled{3} $\nu = 1e-4, T=30, N=10000$; \textcircled{4} $\nu = 1e-5, T=20, N=1000$. The data sets are generated on a $256\times 256$ grid and are subsampled to $64\times 64$.

We see in Table\,\ref{tab:NS} that the proposed MWT Leg outperforms the existing Neural operators as well as other deep NN benchmarks. The MWT models have used a $2$d multiwavelet transform with $k=3$ for the vorticity $w$, and $3$d convolutions in the $A, B, C$ NNs for estimating the time-correlated kernels. The MWT models (both Leg and Chb) are trained for $500$ epochs for all the experiments except for $N=10000, T=30, \nu = 1e-4$ case where the models are trained for $200$ epochs. Note that similar to FNO-2D, a time-recurrent version of the MWT models could also be trained and most likely will improve the resulting $L2$ error for the less data setups like $N=1000, \nu=1e-4$ and $N=1000, \nu=1e-5$. However, in this work we have only experimented with the $3$d convolutions (for $A, B, C$) version.

\begin{table}
\centering
\hspace*{\fill} \\
\hspace*{\fill} \\
\begin{tabular}{p{2cm} p{1.8cm} p{1.8cm} p{1.8cm} p{1.8cm}}
 \hline
 Networks & \makecell[l]{$\nu=1e-3$\\$T=50$\\$N=1000$} &\makecell[l]{$\nu=1e-4$\\$T=30$\\$N=1000$}&\makecell[l]{$\nu=1e-$4\\$T=3$0\\$N=10000$}&\makecell[l]{$\nu=1e-5$\\$T=20$\\$N=1000$}\\
 \hline
 MWT Leg   &\textbf{0.00625}&\textbf{0.1518}&\textbf{0.0667}&\textbf{0.1541}\\
 MWT Chb   & 0.00720 & 0.1574 & 0.0720 & 0.1667\\
 \hline
FNO-3D & 0.0086 &  0.1918 &  0.0820 &  0.1893\\
FNO-2D& 0.0128 &0.1559&0.0973&0.1556\\
U-Net&   0.0245&0.2051&0.1190&0.1982\\
TF-Net &0.0225&0.2253&0.1168&0.2268\\
Res-Net &0.0701&0.2871&0.2311&0.2753\\
 \hline
\end{tabular}
\caption{\label{tab:NS}Navier-Stokes Equation validation at various viscosities $\nu$. Top: Our methods. Bottom: previous works of Neural operators and other deep learning models.}
\end{table}

\subsection{Prediction at higher resolutions}
\label{apssec:highRes}
\begin{figure}[t!]
    \centering
    \includegraphics[width = \linewidth]{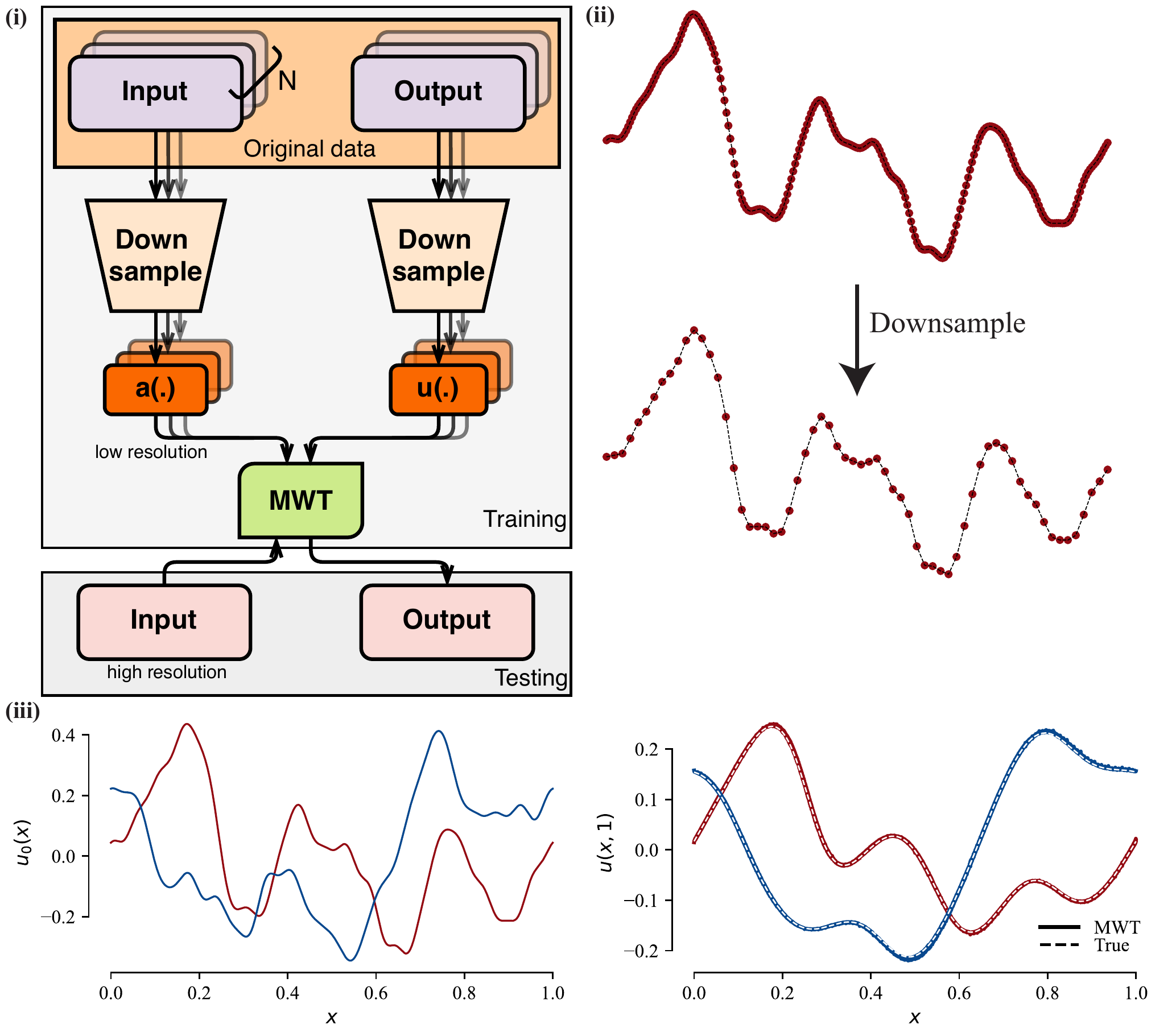}
    \caption{\textbf{Prediction at higher resolution:} The proposed model (MWT) learns the function mapping using the data with a coarse resolution, and can predict the output at a higher resolution. \textbf{(i)} The resolution-extension experiment pipeline. \textbf{(ii)} An example of down-sampling of the associated functions used in the training. \textbf{(iii)} We show two test samples with example-1 marked as blue while example-2 is marked as red. \textbf{Left:} input functions ($u_0$) of the examples. \textbf{Right:} corresponding outputs $u(x,1)$ at $s=8192$ from MWT Leg (trained on $s=256$) of the 2 examples, and their higher-resolution ($s=8192$) ground truth (dotted line).}
    \label{fig:Burger_ex}
\end{figure}

The proposed multiwavelets-based operator learning model is resolution-invariant by design. Upon learning an operator map between the function spaces, the proposed models have the ability to generalize beyond the training resolution. In this Section, we evaluate the resolution extension property of the MWT models using the Burgers' equation dataset as described in the Section\,\ref{ssec:burgers}. A pipeline for the experiment is shown in Figure\,\ref{fig:Burger_ex}. The numerical results for the experiments are shown in Table \ref{tab:test_high}. 
\begin{table}[h]
\centering
\begin{tabular}{p{2cm} p{1.8cm} p{1.8cm} p{1.8cm}}
 \hline
 \diagbox{Train}{Test}& s = 2048&s = 4096&s = 8192\\
 \hline
s=128  &0.0368&0.0389&0.0456\\
s=256 &0.0226&0.0281&0.0321 \\
s=512 & 0.0140 &  0.0191 &  0.0241 \\
 \hline 
\end{tabular}
\caption{\label{tab:test_high} MWT Leg model trained at lower resolutions can predict the output at higher resolutions.}
\end{table}
We see that on training with a lower resolution, for example, $s=256$, the prediction error at 10X higher resolution $s=2048$ is $0.0226$, or $2.26\%$. A sample input/output for learning at $s=256$ while predicting a $s=8192$ resolution is shown in Figure\,\ref{fig:Burger_ex}. Also, learning at an even coarser resolution of $s=128$, the proposed model can predict the output of $2^6$ times the resolution (i.e., $s=8192$) data with an relative $L2$ error of $4.56\%$.
\subsection{Pseudo-Differential Equation}
\label{apssec:pseudoDiff}
\begin{figure}
    \centering
    \includegraphics[width = \linewidth]{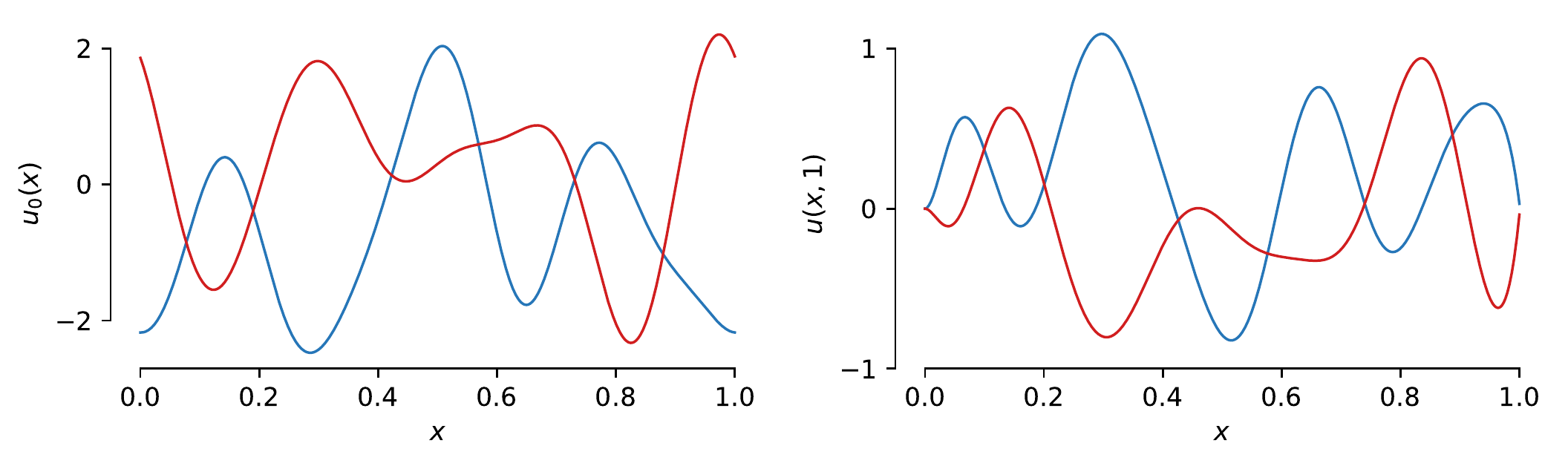}
    \caption{Sample input/output for the PDE of Section\,\ref{apssec:pseudoDiff}. \textbf{Left:} Two input functions ($u_0$) examples in Red and Blue. \textbf{Right:} corresponding outputs ($u(x,1)$) in the same color.}
    \label{fig:EB_03}
\end{figure}

Similar to the experiments presented in the Section\,\ref{ssec:eulerBer} for Euler-Bernoulli equation, we now present an additional result on a different pseudo-differential equation. We modify the Euler-Bernoulli beam to a $3$rd order PDE as follows
\begin{align}
    \begin{aligned}
        \frac{\partial^{3} u}{\partial x^3} - \omega^{2}u &= f(x),\quad \frac{\partial u}{\partial x}\Bigr|_{x=0} = 0, \quad x\in [0,1]\\
        u(0) &= u(1) = 0,\\
    \end{aligned}
    \label{apeqn:3rdOrderPDE}
\end{align}
where $u(x)$ is the Fourier transform of the time-varying displacement, $\omega=215$ is the frequency, $f(x)$ is the external function. The eq. \eqref{apeqn:3rdOrderPDE} is not known to have a  physical meaning like Euler-Bernoulli, however, from a simulation point-of-view it is sufficient to be used as a canonical PDE. A sample force function (input) and 
\begin{wrapfigure}{r}{0.5\linewidth}
\includegraphics[width=\linewidth]{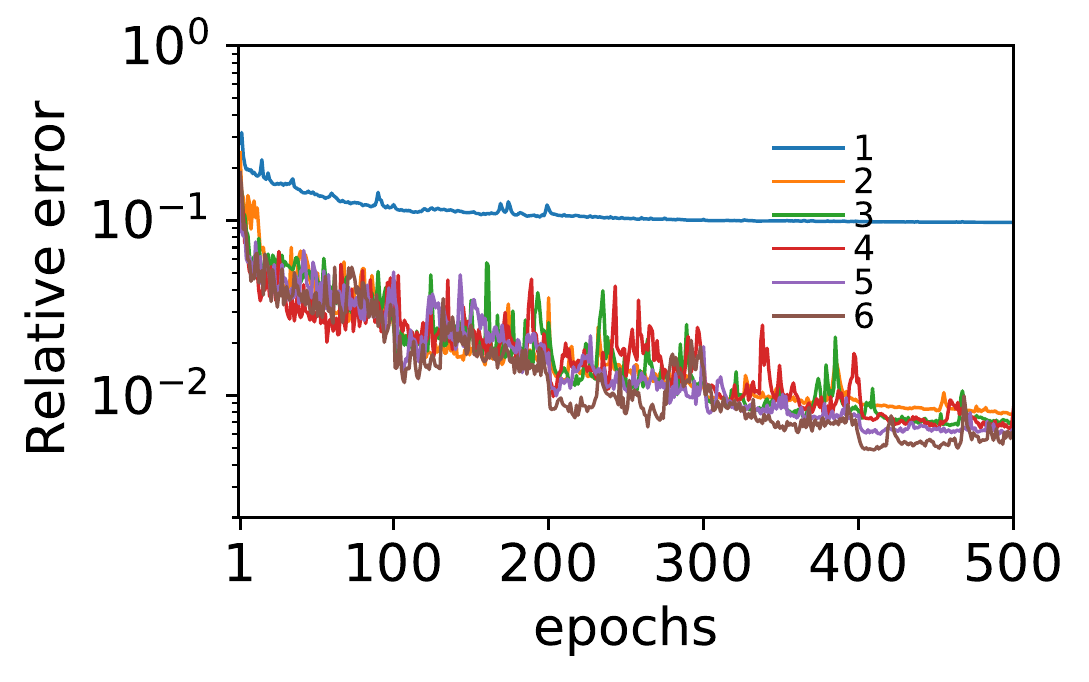}
\caption{Relative $L2$ error vs epochs for MWT Leg with different number of OP basis $k$.}
\label{fig:eulerBernoulli_03}
\vspace{-15pt}
\end{wrapfigure}
the solution of the PDE in \eqref{apeqn:3rdOrderPDE} is shown in Figure\,\ref{fig:EB_03}. The eq. \eqref{apeqn:3rdOrderPDE} is a pseudo-differential equation with the maximum derivative order $T+1=3$. We now take the task of learning the map from $f$ to $u$. In Figure\,\ref{fig:eulerBernoulli_03}, we see that for $k\geq2$, the models relative error across epochs is similar, which again is in accordance with the Property\,\ref{propr:smoothness}, i.e., $k>T-1$ is sufficient for annihilation the kernel away from the diagonal by multiwavelets. We saw a similar pattern for the $4$th order PDE in Section\,\ref{ssec:eulerBer} but for $k\geq 3$. 

\begin{figure}
    \centering
    \includegraphics[width = \linewidth]{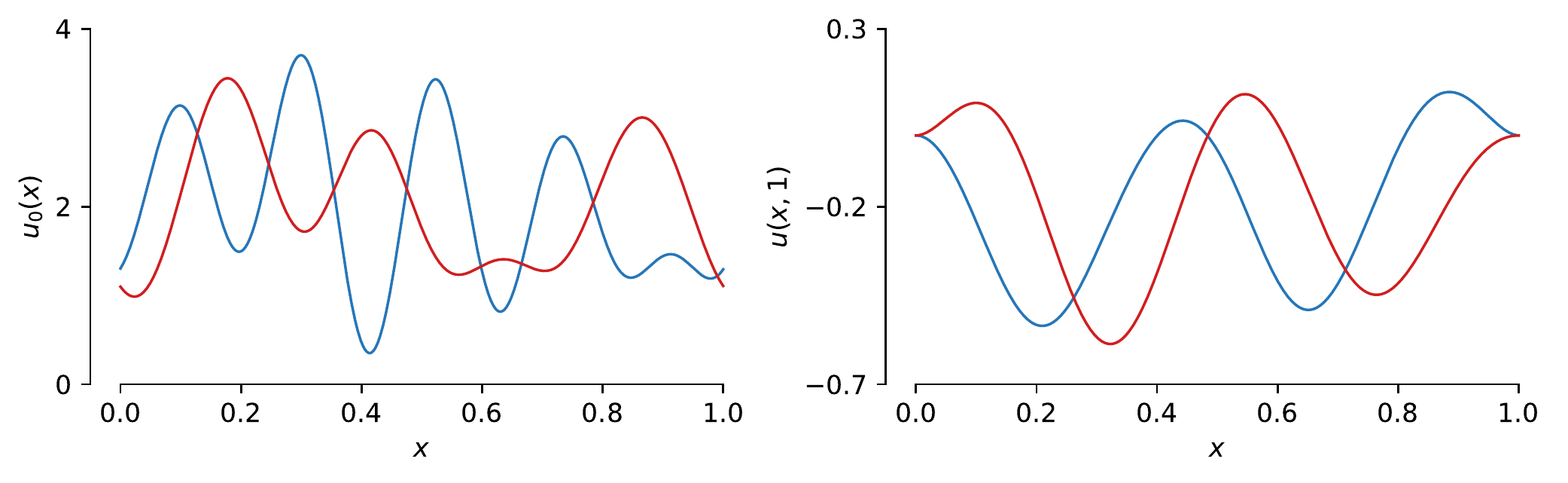}
    \caption{Two examples of $4$th order Euler-Bernoulli equation. \textbf{Left:} Two input functions ($u_0$) in different colors. \textbf{Right:} corresponding outputs ($u(x,1)$) in the same color.}
    \label{fig:EB_04}
\end{figure}

\subsection{Korteweg-de Vries (KdV) Equation}
\label{apsec:kdv}
We present additional results for the KdV equation (see Section\,\ref{ssec:kdv}). First, we demonstrate the operator learning when the input is sampled from a squared exponential kernel. Second, we experiment on the learning behavior of the Neural operators when the train and test samples are generated from different random sampling schemes.
\subsubsection{Squared Exponential Kernel}
\label{apssec:sek}
We sample the input $u_{0}(x)$ from a squared exponential kernel, and solve the KdV equation in a similar setting as mentioned in Section\,\ref{ssec:kdv}. Due to the periodic boundary conditions, a periodic version of the squared exponential kernel \cite{rasmussen2006gaussian} is used as follows.
\begin{equation*}
k(x, x') = \exp\left(-2\frac{\sin^{2}(\pi(x-x')/P)}{L^2}\right),
\end{equation*}
\noindent where, $P$ is the domain length and $L$ is the smoothing parameter of the kernel. The random input function is sampled from $\mathcal{N}(0, K_{m})$ with $K_{m}$ being the kernel matrix by taking $P=1$ (domain length) and $L=0.5$ to avoid the sharp peaks in the sampled function. The results for the Neural operators (similar to Table\,\ref{tab:kdv}) is shown in Table\,\ref{tab:kdvSEK}. We see that MWT models perform better than the existing neural operators at all resolutions.
\begin{table}
\centering
\begin{tabular}{p{2cm} p{1.8cm} p{1.8cm} p{1.8cm} p{1.8cm} p{1.8cm}}
\hline
Networks& s = 64&s = 128&s = 256&s = 512 &s = 1024\\
\hline
MWT Leg    &\textbf{0.00190}&\textbf{0.00214}&\textbf{0.00204}&\textbf{0.00201}&\textbf{0.00211}\\
MWT Cheb &0.00239&0.00241&0.00221&0.00289&0.00348 \\
\hline
FNO & 0.00335 &  0.00330 &  0.00375 &  0.00402 & 0.00456\\
MGNO& 0.0761    &0.0593&0.0724&0.0940&0.0660\\
LNO&   0.0759&0.0756&0.0673 &0.0807& 0.0792\\
GNO &0.0871&0.0801&0.0722&0.0798&0.0777\\
 \hline 
\end{tabular}
\caption{\label{tab:kdvSEK} Korteweg-de Vries (KdV) equation benchmarks for different input resolution $s$ with input $u_{0}(x)$ sampled from a squared exponential kernel. Top: Our methods. Bottom: previous works of Neural operator.}
\end{table}

\subsubsection{Training/Evaluation with different sampling rules}
\label{apssec:diffSample}
The experiments in the current work and also in all of the recent neural operators work \cite{li2020fourier,Li2020MGNO} have used the datasets such that the train and test samples are generated by sampling the input function using the same rule. For example, in KdV, a complete dataset is first generated by randomly sampling the inputs $u_{0}(x)$ from $\mathcal{N}(0, 7^4(\Delta+7^2I)^{-2.5})$ and then splitting the dataset into train/test. This setting is useful when dealing with the systems such that the future evaluation function samples have similar patterns like smoothness, periodicity, presence of peaks. However, from the viewpoint of learning the operator between the function spaces, this is not a general setting. We have seen in Figure\,\ref{fig:kdv_fluc_acc} that upon varying the fluctuation strength in the inputs (both train and test), the performance of the neural operators differ. We now perform an addition experiment in which the neural operator is trained using the samples from a periodic squared exponential kernel (Section\,\ref{apssec:sek}) and evaluated on the samples generated from random fields \cite{Filip2019SmoothRandom} with fluctuation parameter $\lambda$. We see in Table\,\ref{tab:kdvSEKGRF} that instead of different generating rules, the properties like fluctuation strength matters more when it comes to learning the operator map. Evaluation on samples that are generated from a different rule can still work well provided that the fluctuations are of similar nature. It is intuitive that by learning only from the low-frequency signals, the generalization to higher-frequency signals is difficult.
\begin{table}
\centering
\begin{tabular}{p{2cm} p{1.8cm} p{1.8cm} p{1.8cm}}
\hline
Networks& $\lambda = 0.25$&$\lambda = 0.5$&$\lambda = 0.75$\\
\hline
\multicolumn{4}{c}{$s = 64$}\\
\hline
MWT Leg    &0.00819 & 0.00413 &\textbf{0.00202}\\
MWT Cheb   & \textbf{0.00751}        & \textbf{0.00347}         & 0.00210 \\
FNO & 0.0141         &  0.00822        &  0.00404 \\
MGNO        & 0.3701         &0.2030           & 0.0862\\
LNO         &   0.1012       &0.0783           &0.0141\\
\hline
\multicolumn{4}{c}{$s = 256$}\\
\hline
MWT Leg    &0.00690 & \textbf{0.00322} & \textbf{0.00145}\\
MWT Cheb   &\textbf{0.00616}         &0.00344          & 0.00182 \\
FNO& 0.0134          &  0.00901        &  0.00376 \\
MGNO       & 0.4492          &0.2114           & 0.1221\\
LNO        &   0.1306        & 0.0821          & 0.0161\\
\hline
\multicolumn{4}{c}{$s = 1024$}\\
\hline
MWT Leg    &\textbf{0.00641} &0.00408 &\textbf{0.00127}\\
MWT Cheb   &0.00687          & \textbf{0.00333}         & 0.00176 \\
FNO & 0.0141          &  0.00718        &  0.00359 \\
MGNO        & 0.4774          & 0.2805          & 0.1309\\
LNO         &   0.1140        & 0.0752          & 0.0139\\
\hline 
\end{tabular}
\caption{\label{tab:kdvSEKGRF} Neural operators performance when training on random inputs sampled from Squared exponential kernel and testing on samples generated from smooth random functions \cite{Filip2019SmoothRandom} with controllable parameter $\lambda$. The random functions are used as the input $u_{0}(x)$ for Korteweg-de Vries (KdV) equation as mentioned in Section\,\ref{ssec:kdv}. In the test data, $\lambda$  is inversely proportional to sharpness of the fluctuations.}
\end{table}
\makeatletter
\setlength{\@fptop}{0pt} 
\setlength\@fpsep{8pt plus 0.1fil} 
\makeatother
\subsection{Burgers Equation}
\label{apssec:burgersAddit}
The numerical values for the Burgers' equation experiment, as presented in Figure\,\ref{fig:Burgers}, is provided in the Table\,\ref{tab:burgers}.
\begin{table}
\centering
\begin{tabular}{p{2cm} p{1.5cm} p{1.5cm} p{1.5cm} p{1.5cm} p{1.5cm} p{1.5cm}}
 \hline
 Networks& s = 256 &s = 512&s = 1024&s = 2048&s = 4096&s = 8192\\
 \hline
 MWT Leg   &\textbf{0.00199}&\textbf{0.00185}&\textbf{0.00184}&\textbf{0.00186}&\textbf{0.00185}&\textbf{0.00178}\\
 MWT Chb &{0.00402}&{0.00381}&{0.00336}&{0.00395}&{0.00299}&{0.00289}\\
 \hline
FNO  & 0.00332 & 0.00333 & 0.00377 &  0.00346 & 0.00324 &  0.00336\\
 MGNO& 0.0243    &0.0355&0.0374&0.0360&0.0364&0.0364\\
 LNO&   0.0212&0.0221&0.0217&0.0219&0.0200&0.0189\\
 GNO &0.0555&0.0594&0.0651&0.0663&0.0666&0.0699\\
 \hline
\end{tabular}
\caption{\label{tab:burgers}Burgers' Equation validation at various input resolution $s$. Top: Our methods. Bottom: previous works of Neural operators.}
\end{table}

\end{document}